\documentclass{IEEEtran}
\usepackage{svg}
\usepackage{amsmath} 
\usepackage{enumitem}
\usepackage{comment}
\usepackage{amssymb} 
\usepackage{booktabs}
\usepackage{cite}
\usepackage{hyperref}
\AtBeginDocument{%
  }
\begin{document}

\title{RF-PGS: Fully-structured Spatial Wireless Channel Representation with Planar Gaussian Splatting}


\author{Lihao Zhang, \emph{Student Member, IEEE},  Zongtan Li, Haijian Sun, \emph{Senior Member, IEEE} 
\thanks{
L. Zhang and H. Sun (lihao.zhang@uga.edu, hsun@uga.edu) are with the School of Electrical and Computer Engineering, University of Georgia, Athens, GA, 30602 USA.

Z. Li is with the School of Environmental, Civil, Agricultural \& Mechanical Engineering, University of Georgia, Athens, GA, 30602 USA.

Dataset and codes of this paper will be available after paper acceptance at: \url{https://github.com/SunLab-UGA/RF-PGS}
}}

\maketitle


\begin{abstract}
In the 6G era, the demand for higher system throughput and the implementation of emerging 6G technologies require large-scale antenna arrays and accurate spatial channel state information (Spatial-CSI). Traditional channel modeling approaches, such as empirical models, ray tracing, and measurement-based methods, face challenges in spatial resolution, efficiency, and scalability. Radiance field-based methods have emerged as promising alternatives but still suffer from geometric inaccuracy and costly supervision. This paper proposes RF-PGS, a novel framework that reconstructs high-fidelity radio propagation paths from only sparse path loss spectra. By introducing Planar Gaussians as geometry primitives with certain RF-specific optimizations, RF-PGS achieves dense, surface-aligned scene reconstruction in the first geometry training stage. In the subsequent Radio Frequency (RF) training stage, the proposed fully-structured radio radiance, combined with a tailored multi-view loss, accurately models radio propagation behavior. Compared to prior radiance field methods, RF-PGS significantly improves reconstruction accuracy, reduces training costs, and enables efficient representation of wireless channels, offering a practical solution for scalable 6G Spatial-CSI modeling.
\end{abstract}




\section{Introduction}

Sixth-generation (6G) wireless networks are envisioned to provide substantially enhanced system throughput, ultra-low latency, improved reliability, ubiquitous coverage, and advanced sensing capabilities. Achieving these ambitious goals fundamentally relies on the efficient operation of Gigantic Multiple-Input Multiple-Output (MIMO) systems~\cite{bjornson2024gigamimo} and Reconfigurable Intelligent Surfaces (RIS)~\cite{liu2021reconfigurable}, which employ arrays composed of thousands of antenna or reflective elements. These large-scale arrays facilitate extreme spatial reuse of spectrum resources, enable high-resolution environmental sensing, and support reliable communication in high-frequency bands such as millimeter-wave and terahertz~\cite{faisal2020ultramassive}.
The cornerstone of such efficient operation is a comprehensive understanding of the spatial wireless channel, specifically, how radio signals propagate through complex environments to reach the receiver \cite{10599118}. In this regard, traditional pilot-based channel estimation schemes, which model the spatial channel as a channel matrix, are effective for small-scale MIMO but face significant limitations as the array size scales up. The overhead associated with pilot transmission, channel matrix estimation, and beamforming optimization becomes prohibitively high. Moreover, the implicit channel matrix offers limited interpretability for sensing or RIS-related applications.

A more scalable alternative is to bypass the estimation of the full MIMO channel matrix and instead explicitly identify the dominant radio propagation paths between the transmitter (Tx) and receiver (Rx)~\cite{xu2019locationbeamforming}. Minimal resources can then be used to refine or align beams in the angular domain, achieving efficient performance for both communication and sensing. In particular, under generic far-field conditions, radio wave propagation can be approximated by the ray concept. These rays travel along various paths and aggregate at the Rx, forming a deterministic multipath channel model (in contrast to statistical ones such as Rician or Rayleigh fading channels). 
Each specific path, received as a multipath component (MPC), is characterized not only by its path loss and delay, but also by directional and spatial properties such as angle of departure (AoD), angle of arrival (AoA), and the full propagation trajectory. The resulting channel description, termed \emph{Spatial-CSI}~\cite{zhang2024spatial}, extends beyond traditional CSI by incorporating such spatial information. In this paper, we focus on a practical subset of Spatial-CSI: path loss, time-of-flight (ToF), AoD, AoA, and the interaction-induced gain along each path. 

For modeling such deterministic multipath channels, the key insight lies in recognizing that radio propagation is highly dependent on environmental geometry and associated radio frequency (RF) properties. Therefore, accurate modeling is site-specific and emphasizes the connection between the physical environment and the spatial wireless channels embedded within it. Based on this principle, several spatial channel modeling methodologies have been proposed.
First, cluster-based methods, such as those in \cite{GSCM_eval,GSCM_joint_SAGE,GSCM_UAV,GSCM_UAV_3d}, employ the geometry-based spatial-consistent MPC tracking method and the geometric-stochastic channel model specified in 3GPP TR25.996~\cite{3GPP_UMTS_TR25996} and TR38.901~\cite{3GPP_TR38900}.  These models are practical and generalizable, as they infer key reflectors and scatterers solely from online channel estimation without requiring additional modules. As a result, they can be seamlessly integrated into existing wireless systems. However, these models fail to provide the fine-grained spatial detail required by emerging 6G applications. Furthermore, their reliability is constrained by the assumption that both significant scatterers and the moving Rx location can be inferred from online feedback alone, which is often overly idealistic and difficult to realize in practice.

Second, the ray tracing methods~\cite{hoydis2023sionna,3d_raytracingwinert_orekondy2022winert}, initially developed for simulation, are increasingly being considered as a core module in future 6G network architectures. In such architectures, environmental knowledge and Rx location can be integrated through a digital-twin-style backbone. Given this information, ray tracing leverages precise shooting-and-bouncing algorithms within a carefully reconstructed 3D scene. While this method can provide high physical fidelity, its accuracy critically depends on the quality of the reconstructed geometry mesh and the RF properties of surfaces—both of which are challenging to acquire and calibrate in realistic and dynamic environments. Moreover, the high computational complexity of ray tracing severely limits its practicality for real-time applications.

Third, the dataset-based methods~\cite{alkhateeb2019deepmimo,zhang2024wisegrt,zeng2024ckmtutorial,NIST_data} record spatial channels derived from ray tracing simulations or field measurements at discrete 3D voxels, alongside corresponding environmental metadata. These methods offer fast querying and high accuracy, making them attractive for offline analysis and neural network training. However, they are limited by the high cost and scenario-specific nature of data collection, which hinders scalability and generalization. Additionally, such datasets often require considerable storage capacity, especially when targeting high spatial resolution. Without dense sampling, accurate interpolation of directional spatial features (e.g., AoA, AoD) from sparse discrete points becomes significantly more difficult than interpolating scalar quantities like path loss. Furthermore, how to effectively fuse the high-geometry-accuracy ray tracing results with high-fidelity field measurements remains an open and critical research challenge.

\subsection{Radio Radiance Field Methods}

Recently, Radio Radiance Field (RRF) \cite{zhang2024rf} methods have demonstrated strong potential for comprehensive spatial wireless channel modeling. Inspired by visual radiance field techniques originally developed in computer vision for novel view synthesis and scene reconstruction, RRF approaches extend these concepts to the RF domain~\cite{zhang2024rf, zhao2023nerf2, lu2024newrf, yang2025scalable, wen2024wrf, chen2024rfcanvas}. The core idea of RRF methods is to learn a representation that simultaneously reconstructs the scene geometry and models how radio waves radiate from the object points and arrive at the Rx, using either joint visual-RF data or RF data alone. These methods effectively address many of the limitations found in traditional spatial channel modeling approaches.
The pioneering method $\text{NeRF}^2$~\cite{zhao2023nerf2} was the first to employ a neural radiance field to construct a squared RRF indexed by both Tx and Rx positions. This formulation enabled querying of Spatial-CSI at arbitrary locations in the environment. However, it suffers from heavy reliance on densely sampled training data, which makes it difficult to scale in practice.
Subsequently, WRF-GS~\cite{wen2024wrf} was introduced based on the 3D Gaussian Splatting (3DGS) framework, significantly improving training efficiency and query speed. Nonetheless, it relies solely on RF data and lacks the spatial resolution and capability for accurately modeling multi-modal Spatial-CSI.

In parallel, our previous work RF-3DGS~\cite{zhang2024rf} addressed these challenges by introducing a two-stage solution. First, it reconstructs accurate environmental geometry using visual data. Then, using only very sparse RF measurements, it constructs a continuous radio radiance field over this geometry, capturing how radio wave is emitted and reflected by surfaces in the environment. This method also supports accurate encoding of multi-modal Spatial-CSI, including path loss, delay, AoD, and AoA.
Compared to prior RRF approaches, RF-3DGS enables accurate and real-time Spatial-CSI query at arbitrary Rx positions. The learned RRF model is compact in size and supports rapid inference—querying a deterministic multipath channel at a given Rx location takes only a few milliseconds. In terms of reconstruction cost, RF-3DGS requires only a set of normal perspective images and approximately ten minutes for geometry reconstruction. The subsequent RRF construction uses tens of uniformly distributed RF samples and completes under three minutes.

Despite its advantages, RF-3DGS still exhibits several notable limitations.
First, as it is built upon the vanilla 3DGS framework~\cite{kerbl20233d}, originally developed for novel view synthesis, the reconstructed geometry is loosely represented by millions of 3D Gaussian ``ellipsoids." This coarse geometric representation affects the accuracy of the RRF reconstruction and imposes constraints on geometry-dependent downstream tasks.
Second, in modeling multi-modal Spatial-CSI, RF-3DGS introduces a bias toward distance-dependent components. Specifically, path loss is approximated in the logarithmic (dB) scale along outgoing rays from object surfaces, while ToF is not explicitly modeled but rather estimated via a normalized delay approximation. 
Third, the training process heavily relies on the availability of multi-modal Spatial-CSI spectra, namely, path loss, AoD, and delay spectra. However, in realistic scenarios, only the path loss spectrum (e.g., RF ``pictures” that describe the distribution of received signal strength in AoA domain) is readily obtainable. Acquiring accurate AoD and delay spectra remains challenging, which restricts the practical applicability of the method. Expanding this information on the Tx side (e.g., in the AoD domain) requires further transformation introduce additional overhead.
Fourth, the unique characteristics of the RF domain demand more tailored modifications.

\subsection{RF-PGS}

To address the aforementioned limitations of RF-3DGS, we propose Radio Frequency Planar Gaussian Splatting (RF-PGS). As illustrated in Fig.~\ref{fig:3_pipeline_overview}, RF-PGS introduces several key enhancements that improve geometric accuracy, representation fidelity, and training efficiency.
First, RF-PGS replaces the ellipsoidal 3D Gaussians with flattened planar 3D Gaussians as the fundamental geometric primitives. These planar Gaussians are inherently better suited for representing surface geometry. During optimization, the planar Gaussians are explicitly encouraged to align with ground-truth geometry at the cost of a small increase in training time. To further enhance surface smoothness and consistency, we propose a curvature-aware geometry refinement that regularizes the normals of neighboring Gaussians at true surface places, while maintaining the sharp wedge points for modeling diffraction.
With accurately reconstructed scene geometry, RF-PGS enables retrieval of full propagation paths along each outgoing ray within the RRF. This foundation allows for improved modeling of Spatial-CSI. Specifically, for each MPC, the path loss representation can now directly incorporate free-space propagation loss, since the full propagation length is provided. Additionally, ToF can be accurately computed from the geometric distance, eliminating the need for normalized delay approximations used in RF-3DGS.
Importantly, RF-PGS reduces the dependency on hard-to-acquire multi-modal spectra. Training relies solely on the practically available path loss spectra, while AoD is inferred from the retrieved full path geometry, and ToF is computed from distance. This enables direct querying of multi-modal Spatial-CSI in the AoD domain (Tx-side), providing greater flexibility and utility for downstream tasks.
Finally, RF-PGS incorporates RF-domain-specific enhancements. For instance, it explicitly models diffraction by identifying wedge points, which is particularly critical for accurate radio propagation modeling at sub-6~GHz frequencies. Furthermore, a multi-view loss is introduced to mitigate inconsistencies in wireless samples from different viewpoints, improving overall reconstruction robustness.

In the experimental evaluation, RF-PGS demonstrates significant performance improvement over prior state-of-the-art methods. Overall, it achieves superior reconstruction quality across all three evaluation metrics. RF-PGS also excels in the Tx-side spectrum evaluation, indicating that the trained model effectively captures the underlying physics of radio wave propagation. In addition, RF-PGS proves highly capable in supporting a spatial-domain beamforming scenario, accurately providing both the transmitter and receiver with the correct AoD and AoA for dominant paths. This enables high channel gain while minimizing interference to potential nearby users.

\subsection{Contributions.} 
This work introduces RF-PGS, a novel framework for fully-structured spatial wireless channel representation. The main contributions are as follows:
\begin{itemize}
    \item We introduce Planar Gaussian geometry with RF-specific optimizations, enabling accurate radio-surface intersection and curvature estimation for modeling complete radio propagation paths and diffraction effects.
    \item We achieve reconstruction of fully-structured radio radiance field using only Rx-side path loss spectra, along with direct querying at both the Rx and Tx sides, thereby overcoming the major limitations of RF-3DGS.
    \item We design multi-view loss functions to mitigate imperfections introduced during signal processing.
    \item We validate RF-PGS on both simulated datasets (2.4~GHz and 60~GHz) and real-world measurements, demonstrating superior reconstruction quality, generalizability, and efficiency.
\end{itemize}

\section{Preliminaries of RRF Reconstruction}
Before presenting our proposed method, RF-PGS, we first provide a brief overview of how an RRF represents the deterministic multipath channel at arbitrary Rx positions. We also shortly revisit how our previous method, RF-3DGS, facilitates a more practical RRF reconstruction pipeline compared to prior approaches.

\begin{figure}[h]
    \centering
    \includegraphics[width=0.45\textwidth]{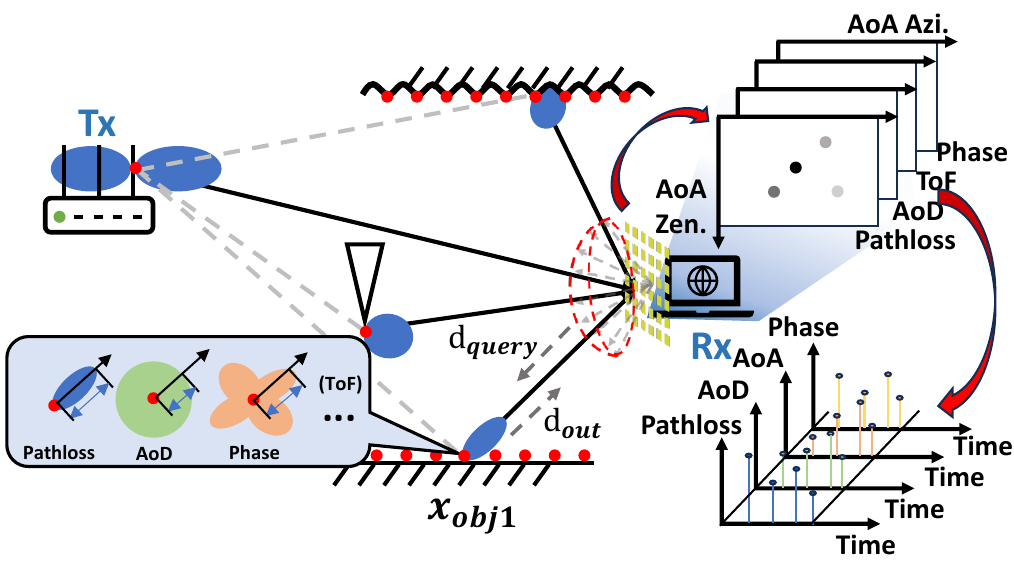} 
    \vspace{-4mm}
    \caption{Illustration of the RRF representation. The four dominant radio propagation paths are represented by the RRF and received as four MPCs at the Rx, which appear as four distinct pixels in the multi-channel spectrum or as four impulses in the CIR.}
    \vspace{-6mm}
    \label{fig:1_RRF_concept_pgs}  
\end{figure}
\vspace{-6mm}

\subsection{Acquiring an Arbitrary multipath Channel from the RRF Representation}
In RF-3DGS, we formally defined a practical RRF representation as $\mathcal{R}(\mathcal{E}, \mathcal{T}) = \left\{ \mathbf{c}(\mathbf{x}_{\text{obj}}, \mathbf{d}),\ \alpha(\mathbf{x}) \right\}$. This RRF representation $\mathcal{R}$ maps a wireless communication scenario—characterized by the environment $\mathcal{E}$ and the transmitter configuration $\mathcal{T}$—to two core functions: the vector-valued radio radiance field function $\mathbf{c}(\mathbf{x}_{\text{obj}}, \mathbf{d})$ and the geometry density field function $\alpha(\mathbf{x})$.
The RRF function $\mathbf{c}(\mathbf{x}_{\text{obj}}, \mathbf{d})$ is defined over all continuous 3D object positions $\mathbf{x}_{\text{obj}}$, and describes the radio radiance emitted in all continuous directions $\mathbf{d}$ from each object position (as only object points can emit radiance, unlike free space where waves merely propagate through).
The density field $\alpha(\mathbf{x})$ encodes the scene geometry by assigning a scalar value to each point in 3D space. It takes a value of zero in free space, low values for translucent materials, and high values for solid objects.

The concept of the RRF function is difficult to visualize directly, but its behavior becomes more intuitive when examined under local conditions. For instance, consider a particular object point within the RRF representation, such as $\mathbf{x}_{\text{obj1}}$, highlighted by the blue callout box in Fig.~\ref{fig:1_RRF_concept_pgs}. At this point, the RRF function $\mathbf{c}(\mathbf{x}, \mathbf{d})$ degrades to a vector-valued spherical function, which we denote as $\mathbf{c}_{\mathbf{x}_{\text{obj1}}}(\mathbf{d})$.
Each dimension of $\mathbf{c}_{\mathbf{x}_{\text{obj1}}}(\mathbf{d})$ corresponds to the distribution of a specific Spatial-CSI component (e.g., path loss, AoD, ToF, phase shift, etc.) across all outgoing directions $\mathbf{d}$, as illustrated in Fig.~\ref{fig:1_RRF_concept_pgs}. When a Rx lies within the line-of-sight (LoS) of this object point, the outgoing direction toward the Rx is denoted as $\mathbf{d}_{\text{out}}$. The radiance along this $\mathbf{d}_{\text{out}}$ is assumed to contribute to an MPC received at the Rx, and its Spatial-CSI vector can then be extracted as $\mathbf{c}_{\mathbf{x}_{\text{obj1}}, \mathbf{d}_{\text{out}}}$.
At the Rx side, this Spatial-CSI vector forms a ``pixel” in the multi-channel radio spatial spectrum. The azimuth and zenith AoA of this MPC, derived from $-\mathbf{d}_{\text{out}}$, determine the 2D pixel index, while the values of the Spatial-CSI components define the pixel intensities across different spectrum channels.
As illustrated in Fig.~\ref{fig:1_RRF_concept_pgs}, the four dominant MPCs correspond to four distinct pixels in the multi-dimensional spectrum, while weaker diffuse MPCs are omitted.

To query the complete deterministic multipath channel at an arbitrary Rx from the RRF representation, a practical strategy is to uniformly query a set of discrete directions $\mathbf{d}_{\text{query}}$ within the Rx’s field of view (FoV)—that is, the angular range over which the Rx maintains a valid antenna gain. These directions point from the Rx toward the entire environment to capture all possible incoming MPCs.
For each query direction $\mathbf{d}_{\text{query}}$, a ray is cast from the Rx to identify the first intersected object point $\mathbf{x}_{\text{obj}}$, which highlights the necessity of including geometric information within the RRF representation. If there exists a strong MPC arriving at the Rx from this first intersected $\mathbf{x}_{\text{obj}}$,  the Spatial-CSI value of this MPC can be retrieved as $\mathbf{c}(\mathbf{x}_{\text{obj}}, -\mathbf{d}_{\text{query}})$.
By aggregating the results from all query directions, we construct the multi-channel radio spatial spectrum at the Rx. This spectrum can further be transformed into the traditional multipath channel representation in the form of a channel impulse response (CIR). The primary distinction is that the radio spatial spectrum unfolds the Spatial-CSI across the AoA domain, while the CIR unfolds it over the time domain.
Since this equivalence holds for any Rx position, we can conclude that the RRF representation encodes the complete set of possible multipath channels within the given wireless communication scenario $(\mathcal{E}, \mathcal{T})$. A practical RRF method should leverage carefully designed representation structure, querying approach, and reconstruction strategy to enable this mapping from $(\mathcal{E}, \mathcal{T})$ to the two continuous functions $\mathbf{c}(\mathbf{x}_{\text{obj}}, \mathbf{d})$ and $\alpha(\mathbf{x})$.

\begin{figure}[h]
    \vspace{-4mm}
    \centering
    \includegraphics[width=0.45\textwidth]{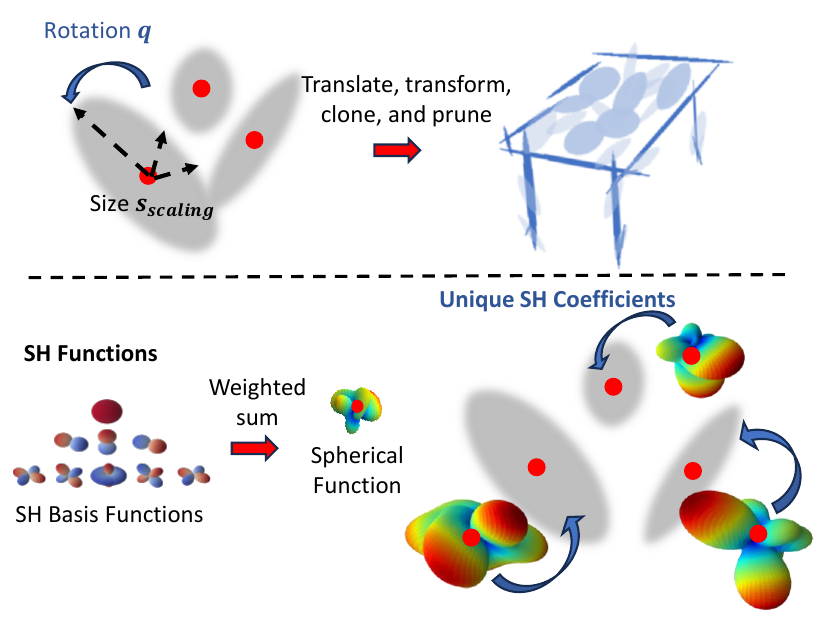} 
    \vspace{-2mm}
    \caption{RF-3DGS representation structure.}
    \label{fig:2_RRFwithRF3dgs}  
\end{figure}

\begin{figure*}[h]
    \centering
    \includegraphics[width=\textwidth]{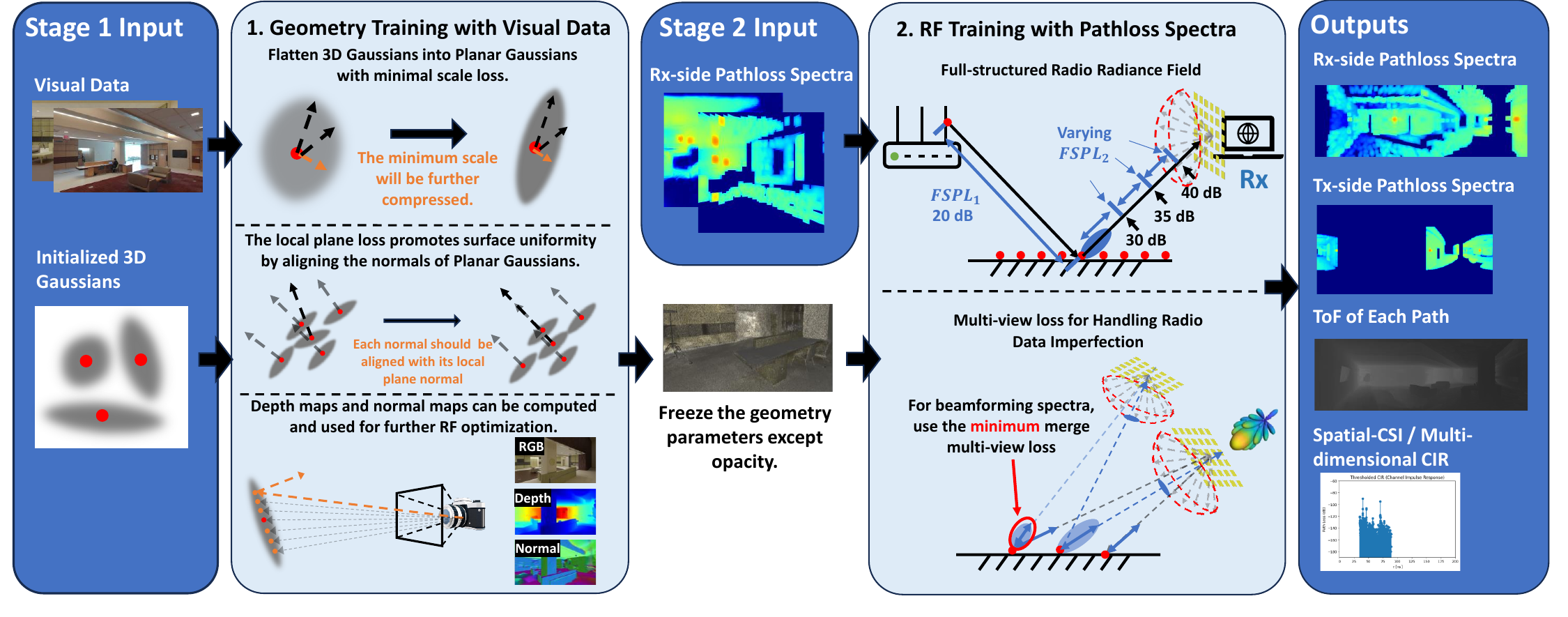} 
    \caption{RF-PGS pipeline.}
    \label{fig:3_pipeline_overview}  
    \vspace{-5mm}
\end{figure*}
\vspace{-5mm}

\subsection{RRF Reconstruction with RF-3DGS}

In RF-3DGS, we adopt an explicit and efficient representation to model both the geometry density field $\alpha(\mathbf{x})$ and the RRF function $\mathbf{c}(\mathbf{x}_{\text{obj}}, \mathbf{d})$. 
As illustrated in Fig.~\ref{fig:2_RRFwithRF3dgs}, the density field is represented as an aggregate of millions of 3D Gaussian distributions. Each Gaussian is parameterized by learnable geometric attributes, including a base density $\alpha_g$, a center position (mean) $\mathbf{x}$, and a shape defined by a scaling vector $\mathbf{s}_{\text{scale}}$ and a rotation quaternion $\mathbf{q}$. During training, these parameters are optimized to deform and align the Gaussians with the ground-truth scene geometry.
The RRF function $\mathbf{c}(\mathbf{x}_{\text{obj}}, \mathbf{d})$ is defined at the level of each Gaussian, as radiance is assumed to be emitted from the Gaussians themselves. In particular, for each Gaussian, its radiance is represented using Spherical Harmonics (SH), which approximate spherical functions via weighted combinations of orthogonal basis functions. Consequently, each Gaussian is associated with multiple sets of SH coefficients, where each set models the directional distribution of a specific Spatial-CSI component.
Importantly, the SH function associated with a Gaussian does not describe radiance emitted from its center alone. Instead, it captures the uniform but view-dependent ``color” (Spatial-CSI) of the entire Gaussian ellipsoid across all directions.
Thus, for any spatial object point $\mathbf{x}_{\text{obj}}$, the corresponding radiance $\mathbf{c}_{\mathbf{x}_{\text{obj}}}(\mathbf{d})$ is computed as a density-weighted combination of the SH functions from all Gaussians that overlap with $\mathbf{x}_{\text{obj}}$—i.e., those whose densities at this point are non-negligible. While this representation structure is complicated, it enables highly efficient querying and accurate reconstruction.

To query the Spatial-CSI at a specific Rx position from the RF-3DGS representation, we still query a set of directions within the Rx's FoV, under the assumption that MPCs may arrive along all these directions. This constitutes the query model, the organization of the query directions, which is application-dependent. 
For general-purpose modeling, it is preferable to align the query directions with the pixel grid of a panoramic camera, using an equirectangular projection. In contrast, for sensing-related tasks, a perspective camera model (i.e., pinhole projection) is favored, as it better supports the use of computer vision techniques during downstream processing.
Given that millions of 3D Gaussians collectively define a continuous density field, and each is associated with an SH-based radiance function, RF-3DGS uses a parallelized splatting strategy. In this approach, each Gaussian ellipsoid is projected onto the Rx’s angular domain as a 2D ellipse, from near to far, such that all query directions intersecting this ellipse can simultaneously accumulate the Gaussian's contributions to their final Spatial-CSI values.

However, when focusing on a single query direction, we can still interpret the parallelized process as launching a query ray into the RRF representation. The contribution of Gaussians along this ray is aggregated using an alpha-blending formulation that models two key physical principles: (1) only Gaussians with high density significantly affect the query result, and (2) nearer Gaussians should occlude farther ones depending on their density.
For a receiver located at $\mathbf{x}_{\text{Rx}}$ and a query direction $\mathbf{d}_{\text{query}}$, the received Spatial-CSI value along that ray, denoted as $\mathbf{m}_{\text{recv}}(\mathbf{x}_{\text{Rx}}, \mathbf{d}_{\text{query}})$, is computed as:
\begin{equation}
    \mathbf{m}_{\text{recv}}(\mathbf{x}_{\text{Rx}}, \mathbf{d}_{\text{query}}) = \sum_{k=1}^{K} T_k \cdot \alpha_k \cdot \mathbf{c}_k(-\mathbf{d}_{\text{query}}),
    \label{eq:alpha-blending}
\end{equation}
\begin{equation}
    T_k = \exp\left(-\sum_{j=1}^{k-1} \alpha_j\right),
\end{equation}
where $\alpha_k$ denotes the density of the $k$-th Gaussian intersected by the ray, $T_k$ is the accumulated transmittance of all preceding Gaussians, and $\mathbf{c}_k(-\mathbf{d}_{\text{query}})$ is the SH-evaluated Spatial-CSI value contributed by the $k$-th Gaussian along the query direction.

Given the representation structure and the differentiable querying method, the reconstruction process in RF-3DGS is straightforward: it infers the radio spatial spectra at the training poses and minimizes the loss between the queried results and the ground truth. This loss is then backpropagated through the differentiable querying pipeline to update the learnable parameters of the representation—namely, the geometric attributes of the 3D Gaussians and the coefficients of their associated SH functions.
The success of this reconstruction hinges on multi-view supervision, which ensures that the Gaussians are translated and transformed to match the true scene geometry, and that their SH functions collaboratively approximate the underlying RRF. However, this process demands high-resolution, low-distortion training data, which is challenging to obtain for radio spatial spectra due to hardware and environmental limitations.
To address this, RF-3DGS leverages low-cost, high-quality visual data to first train the geometric structure of the Gaussians. Once a reliable geometry is established, RF data is used to optimize the SH functions that encode directional radiance. This cross-modal training strategy enables low-quality directional radio signals to be accurately aligned with precise scene geometry, leading to significantly improved reconstruction quality compared to the raw RF input.
Further details on the theoretical foundation of RRF and the RF-3DGS method can be found in~\cite{zhang2024rf}.

\begin{figure}[h]
    \vspace{-2mm}
    \centering
    \includegraphics[width=0.45\textwidth]{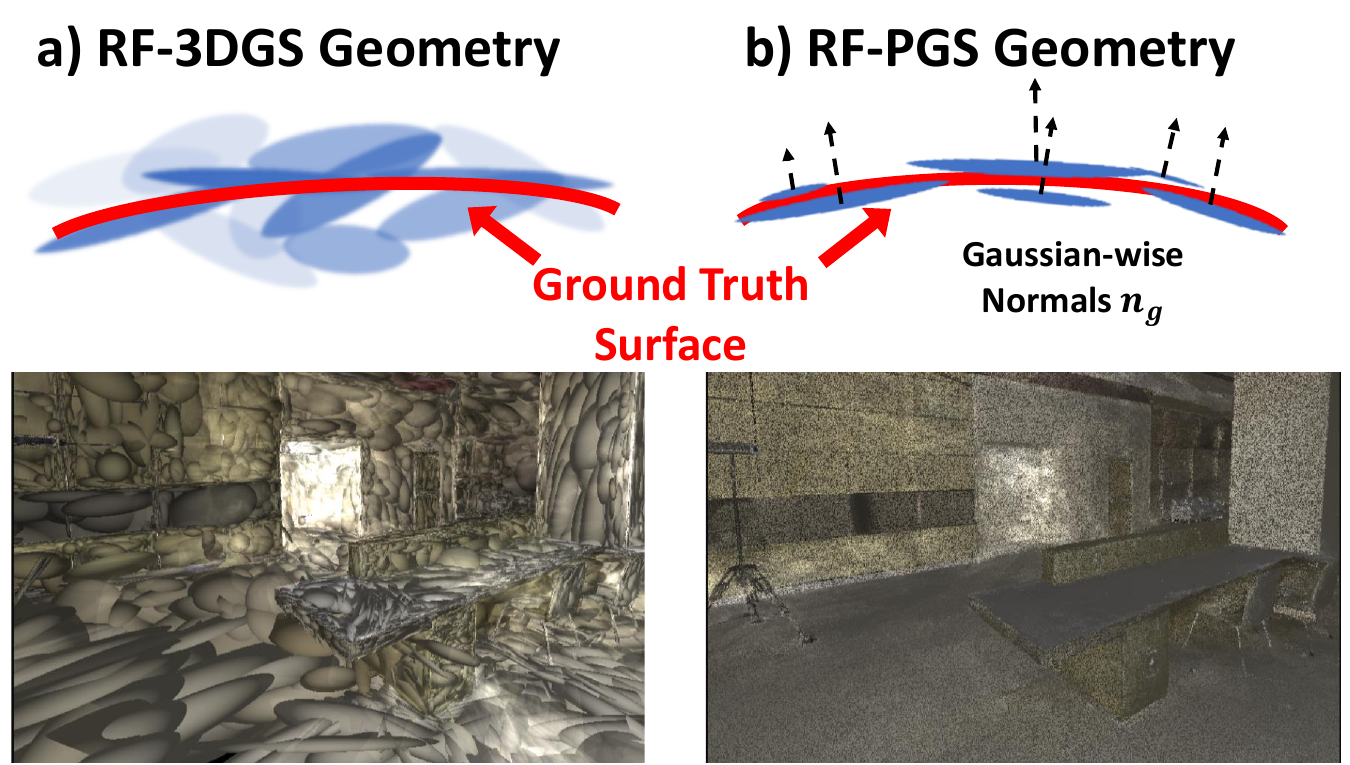} 
    \vspace{-2mm}
    \caption{Planar Gaussian geometry. Each Gaussian primitive is represented as a tiny thin disk with an associated normal $\mathbf{n}_g$.}

    \label{fig:4_direct_geometry_comp}  
    \vspace{-4mm}
\end{figure}

\section{RF-PGS}
Although RF-3DGS outperforms previous methods, it still presents certain limitations, as noted in the Introduction. To address these challenges, we propose RF-PGS.
First, RF-PGS introduces a more accurate geometric foundation for subsequent optimization stages, enabled by the use of dense Planar Gaussians and a curvature-aware surface uniformity loss that encourages geometric consistency across the scene.
Second, RF-PGS supports unbiased modeling of key Spatial-CSI components, including path loss, ToF, AoD, and AoA for each MPC through accurate estimation of ray-surface intersections.
Third, the method significantly reduces the data requirement during training: only the path loss spectra are needed. This enables practical deployment, as accurate AoD and delay spectra are typically difficult to obtain. Moreover, the inference process at the Tx-side is simplified, avoiding the need for domain transformations from AoA to AoD.
Finally, to better account for the unique characteristics of the RF domain, RF-PGS incorporates RF-specific adaptations. These include explicitly modeling wedge diffractors in the environment to improve diffraction modeling and enhance generalization in downstream tasks, as well as introducing two types of multi-view consistency losses to mitigate the inevitable inconsistencies in measured radio spatial spectra.

\vspace{-3mm}

\subsection{Optimizing Geometry for Radio Radiance Field}

When representing geometry for an RRF, there are three characteristics that differ significantly from those in CV applications.
First, all geometric primitives must lie very close to the true object surfaces. This proximity ensures that the reconstructed radio radiance can be correctly mapped to its actual physical position. In contrast, CV-based methods do not strictly enforce this requirement, as novel view synthesis only focuses on the visual fidelity of the final rendering result which is the task that the vanilla 3DGS designed for~\cite{kerbl20233d}. Even in 3D reconstruction tasks, geometry is often extracted through post-processing, rather than requiring primitives to adhere strictly to surface boundaries during primary training.
Second, the improved RRF reconstruction process requires efficient and accurate identification of radio-object interactions. In practice, this means finding the radio-surface intersection point of a query ray and the geometry representation. Once the intersection is determined, the corresponding world coordinate can be directly computed for further optimization.
Third, it is crucial to distinguish between surface points and wedge points. While this distinction is not critical in the CV domain, it is essential for RF-related downstream tasks—such as performing detailed diffraction modeling at wedge points and extracting RF-parameters on flat surfaces. 

In RF-3DGS, these three aspects remain insufficiently addressed. As shown in Fig.~\ref{fig:4_direct_geometry_comp}(a), the method uses loosely aligned, translucent 3D Gaussians as geometric primitives. While this design improves training efficiency, it yields inaccurate geometry for the subsequent RF training stage. Consequently, the imperfect radio data can only be coarsely mapped onto this loose geometry, leading to visible floating artifacts around object boundaries in the reconstructed radio radiance field. Moreover, the lack of precise geometry makes accurate and efficient estimation of radio–surface intersections infeasible, limiting the potential for further RF-specific optimizations.

\subsubsection{Use of Flattened Planar Gaussians as Primitives}
Since geometric primitives form the foundation of radio field reconstruction, we begin by enhancing this aspect by introducing a denser and more accurate surface representation. Prior geometry-focused extensions of 3DGS~\cite{guedon2024sugar,huang20242_2dgs,chen2024pgsr} show that replacing 3D Gaussians with flattened or 2D Gaussians can significantly improve geometric accuracy without sacrificing training or rendering efficiency, while retaining the flexibility of primitive-based representations. In contrast, Signed Distance Function (SDF)-aided approaches~\cite{yu2024gsdf, lyu20243dgsr, azinovic2022neural}, which offer high fidelity but are less suitable for RRF reconstruction.
Therefore, in RF-PGS, we adopt Planar Gaussians as primitives, as illustrated in Fig.~\ref{fig:4_direct_geometry_comp}(b). These thin disks encourage the optimizer to align them more precisely, avoiding the loose distribution often seen with original 3D Gaussians.

Planar Gaussians are formed by flattening 3D Gaussians through a minimum-scale loss term, defined as:
\begin{equation}
    L_{\text{scale}} = \left\| \min(s_1, s_2, s_3) \right\|_1,
\end{equation}
where $s_1$, $s_2$, and $s_3$ are the components of the scaling vector $\mathbf{s}_{\text{scale}}$, which controls the spread of the Gaussian along three orthogonal directions in its local coordinate system, as illustrated in Fig.~\ref{fig:3_pipeline_overview}. This loss compresses the smallest dimension, collapsing each primitive into a thin disk. As shown in Fig.~\ref{fig:4_direct_geometry_comp}(b), this yields a denser and more explicit surface representation, enabling accurate mapping of radio radiance to physical locations and greatly reducing the floating artifacts (Fig.~\ref{fig:8_RF-PGS_rendering}(c)). The surface normal $\mathbf{n}_g$ of each Planar Gaussian in world coordinates can then be derived from its rotation quaternion $\mathbf{q}$ and the axis of minimum scale.

\begin{figure}[h]
    \centering
    \includegraphics[width=0.45\textwidth]{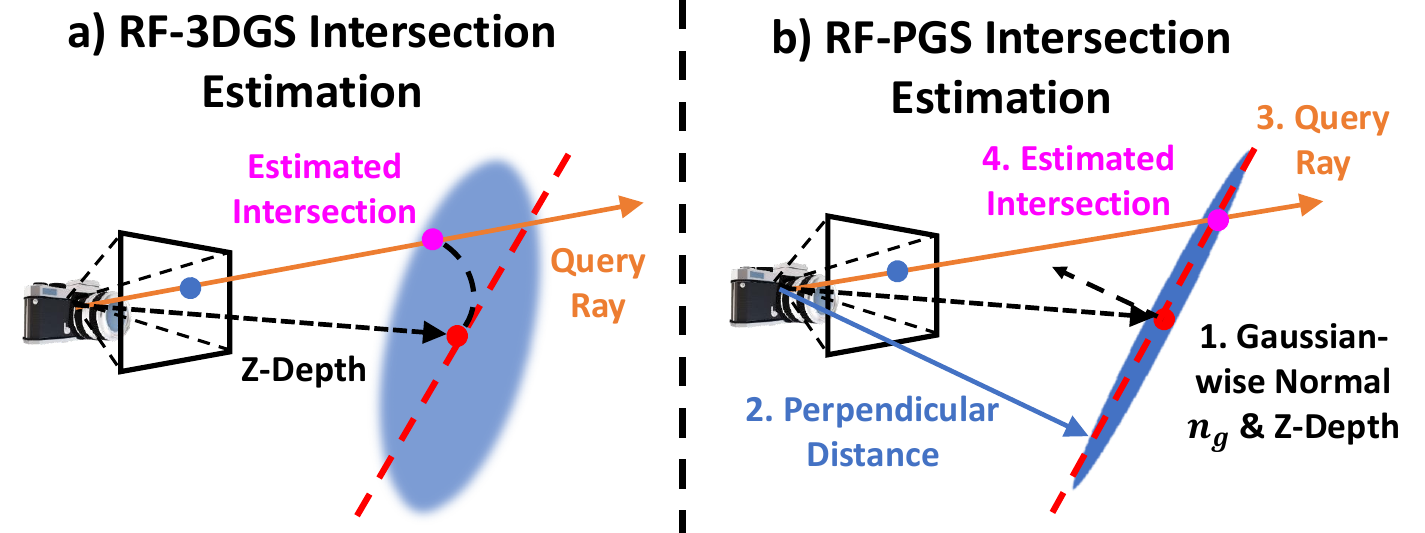} 
    \vspace{-2mm}
    \caption{Estimate Radio-Surface Intersections.}
    \label{fig:5_intersection_estimation}  
    \vspace{-2mm}
\end{figure}

\subsubsection{Estimating Radio–Surface Intersections}

Accurate object surface geometry plays a crucial role in RRF reconstruction. In particular, identifying the intersection between the radio ray and the surface amounts to estimating the depth from the array (or camera for the first stage) center to the true surface along each query ray.
To estimate this depth, we adopt a method analogous to the alpha-blending process used in radio radiance accumulation. Specifically, we compute an aggregated depth by alpha-blending the individual depths to each Gaussian, weighted by their density and accumulated transmittance along the ray. However, in RF-3DGS, a key limitation is that only a Gaussian-wise z-depth is available, which is defined as the Euclidean distance from the camera center to the Gaussian center. This approximation introduces significant errors, especially when the Gaussian center does not lie on the query ray.
As illustrated in Fig.~\ref{fig:5_intersection_estimation}(a), consider the orange query ray from the array center to a Gaussian whose center is offset from the query ray. Using z-depth alone to estimate the intersection leads to large deviation from the true intersection point between the query ray and the surface represented by the Gaussian. The spherical shape and loose spatial distribution of 3D Gaussians make accurate depth calibration infeasible.

\begin{figure}[h]
    \vspace{-3mm}
    \centering
    \includegraphics[width=0.45\textwidth]{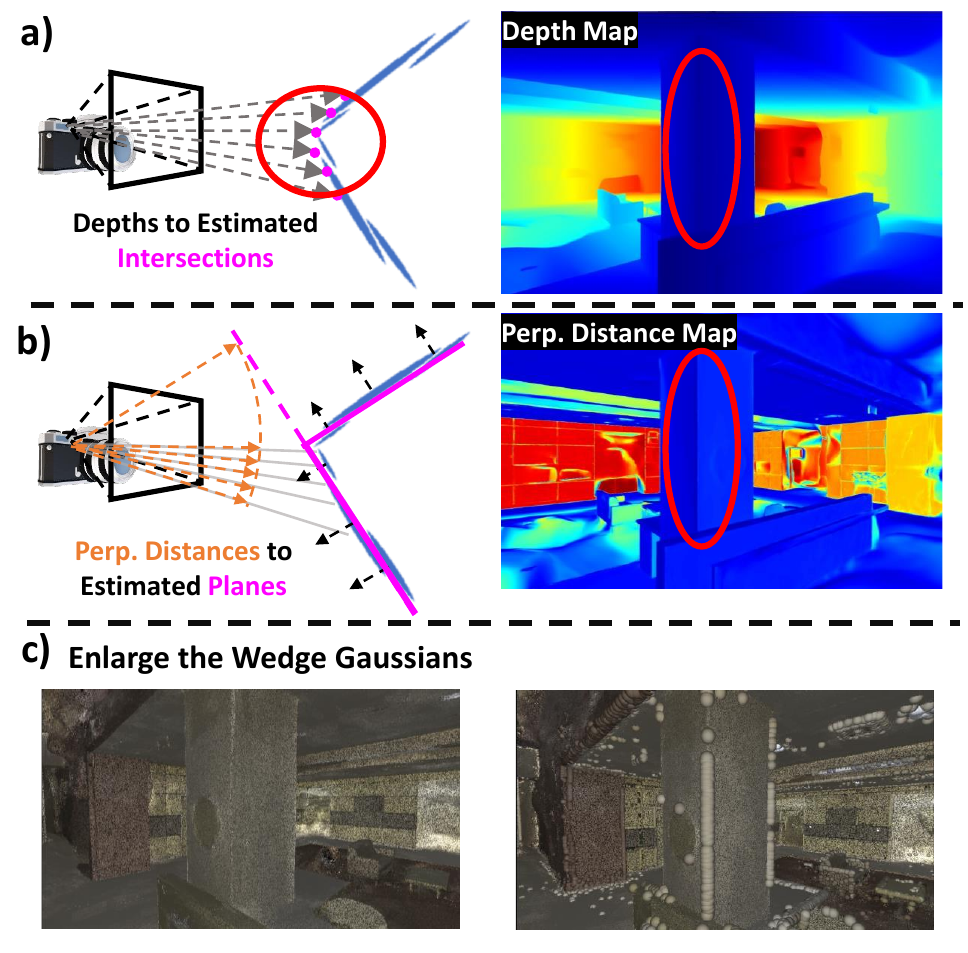} 
    \vspace{-3mm}
    \caption{Curvature-aware Geometry Refinement.}
    \vspace{-3mm}
    \label{fig:6_curverture_estimation}  
\end{figure}

In contrast, Planar Gaussians provide both a surface-aligned shape and an explicit Gaussian-wise normal, making geometric intersection calibration straightforward. As shown in Fig.~\ref{fig:5_intersection_estimation}(b), given a Planar Gaussian with center $\mathbf{x}_g$, normal $\mathbf{n}_g$, and a query direction $\mathbf{d}_{\text{query}}$ from the array or camera center $\mathbf{x}_{\text{Rx}}$, the perpendicular distance from the ray origin to the Gaussian plane is calculated as:
\begin{equation}
    d_{\text{perp}} = \left| (\mathbf{x}_{\text{Rx}} - \mathbf{x}_g) \cdot \mathbf{n}_g \right|.
\end{equation}
Using this, we compute the plane depth, which is the distance from $\mathbf{x}_{\text{Rx}}$ to the estimated intersection point along each query ray:
\begin{equation}
    d_{\text{depth}} = \left| \frac{d_{\text{perp}}}{\mathbf{d}_{\text{query}} \cdot \mathbf{n}_g} \right|.
\end{equation}
This Gaussian-wise plane depth is then used in place of $\mathbf{c}_k(\mathbf{d}_{\text{query}})$ in Eq.~\ref{eq:alpha-blending} to compute an aggregated plane depth along the query ray. This leads to the estimate of the intersection point in the first dense region encountered by the ray, consistent with physical intuition and ray-tracing principles.
Across all query rays, these aggregated plane depths form a depth map,  as shown in Fig.~\ref{fig:6_curverture_estimation}(a). Furthermore, the perpendicular distances and Gaussian-wise normals can also be alpha-blended across the rays to generate a perpendicular distance map ( Fig.~\ref{fig:6_curverture_estimation}(b)) and a normal map (Fig.~\ref{fig:7_local_plane_normal}). These maps are essential for the subsequent two enhancements introduced in RF-PGS.

\subsubsection{Curvature-Aware Geometry Refinement}

To enable further RF-specific geometric refinement, it is essential to account for the curvature of the object surface. In particular, sharp wedge structures should be identified and emphasized to more accurately represent continuous diffraction effects, while flat surface regions should be refined to improve uniformity. To this end, we introduce a novel curvature estimation method applied in the later training iterations, which allows us to distinguish wedge points from smooth surfaces and to align neighboring Planar Gaussians based on local curvature-aware surface normals.
Our curvature estimation primarily leverages two maps generated from the query rays during the final stages of training (after approximately 27,000 iterations, out of 30,000 in our implementation), when the geometry is already reasonably accurate but still lacks fine-level refinement. The first is the plane depth map, shown in Fig.~\ref{fig:6_curverture_estimation}(a), which records the estimated distance from the array or camera center to the intersection point along each query ray. The second is the perpendicular distance map, shown in Fig.~\ref{fig:6_curverture_estimation}(b), which encodes the distance from the ray origin to the intersection plane along the surface normal.
On flat surfaces, the perpendicular distance map remains relatively uniform (i.e., has low image gradients), while the plane depth map varies smoothly due to perspective effects. In contrast, wedge points exhibit strong gradients in the perpendicular distance map but still relatively low gradients in the depth map. We exploit this contrast to estimate curvature: true high-curvature regions are thus characterized by strong gradients in the perpendicular distance map while maintaining low gradients in the depth map. This criterion effectively differentiates sharp features from smooth but spatially varying surfaces.

For instance, 2D edges in images between the slightly curved structures (e.g., cylinders or rounded furniture) and distant background regions can also produce high gradients in the perpendicular map. However, these 2D edges typically coincide with strong gradients in the depth map due to background separation, failing to satisfy the proposed criteria and thus preventing their misclassification as high-curvature wedge points. In contrast, true 3D wedges appear consistently across multiple sampled training views and satisfy both gradient conditions, enabling robust identification of such high-curvature regions.  
Once such wedge points are identified, we modify the corresponding Planar Gaussians, specifically those with the largest contributions along the radio rays intersecting these wedge regions. Their shapes are enlarged into rounded forms, as illustrated in Fig.~\ref{fig:6_curverture_estimation}(c). This allows each high-curvature region to be represented by a single, enlarged Planar Gaussian whose SH function captures the entire continuous diffraction pattern, rather than relying on multiple small, fragmented Gaussians on either side of the wedge.  
This strategy is particularly effective given the relatively low resolution of later used radio spatial spectra. A diffraction MPC originating from such a wedge point, which spreads across many pixels in the Rx AoA domain, can be coherently aggregated into the SH representation of such a single enlarged wedge Gaussian. This improves both the fidelity of diffraction modeling and the utility of the representation for downstream tasks.  

\begin{figure}[h]
    \centering
    \includegraphics[width=0.47\textwidth]{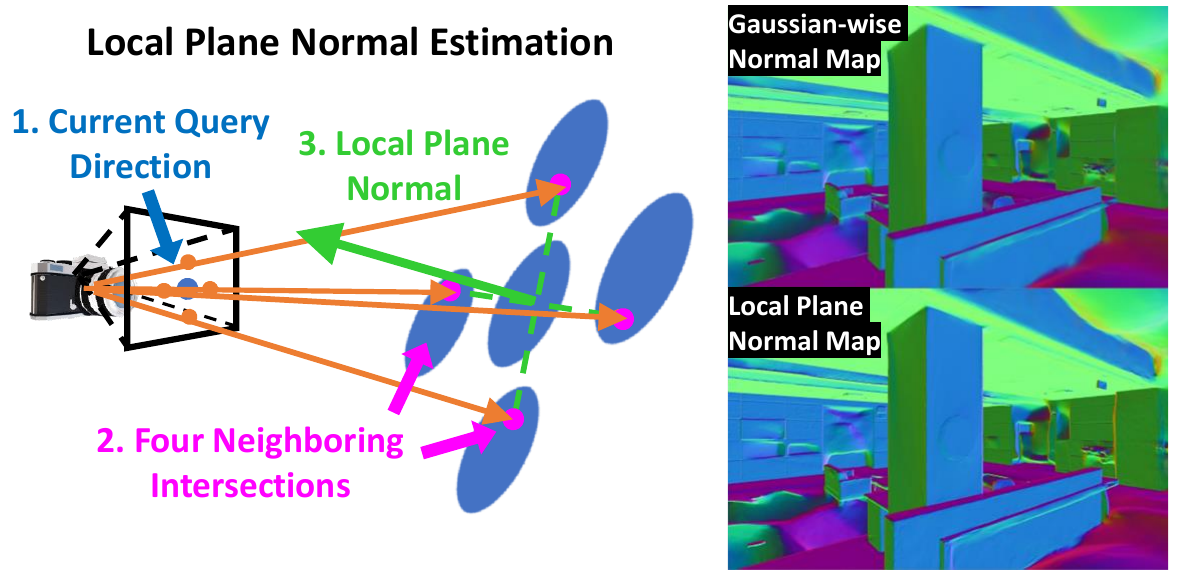} 
    \vspace{-3mm}
    \caption{Local plane normal alignment.}
    \vspace{-5mm}
    \label{fig:7_local_plane_normal}  
\end{figure}

With the estimated curvature, we further improve surface uniformity by minimizing the angular discrepancy between each Gaussian’s normal and its corresponding local plane normal. The local plane normal is estimated from the intersections of neighboring query rays and serves as a reliable reference for enforcing geometric consistency across adjacent Planar Gaussians.  
As illustrated in Fig.~\ref{fig:7_local_plane_normal}, to estimate the local plane normal at a given query direction (indicated in blue), we first select its four neighboring pixel directions in the angular domain and retrieve their corresponding intersection points. Using the current view’s projection matrix, these intersections are projected into 3D coordinates $\{\mathbf{x}_{\text{nb}_1}, \dots, \mathbf{x}_{\text{nb}_4}\}$, indicated as purple points in the figure. The local plane normal $\mathbf{N}_{\text{local}}$ is then computed as
\begin{equation}
    \mathbf{N}_{\text{local}} = (\mathbf{x}_{\text{nb}_1} - \mathbf{x}_{\text{nb}_3}) \times (\mathbf{x}_{\text{nb}_2} - \mathbf{x}_{\text{nb}_4}),
\end{equation}
which corresponds geometrically to the cross product of two diagonal vectors formed by connecting opposing neighbors. This yields the direction perpendicular to the local surface patch, oriented toward the camera.  
To enforce consistency, we define a surface uniformity loss that penalizes the angular difference between the blended Gaussian-wise normal and the estimated local plane normal. This loss is weighted by the previously estimated curvature and a predefined coefficient, ensuring that the refinement is emphasized on smooth surface regions while preserving high-curvature structures such as wedge points.  
As shown in Fig.~\ref{fig:7_local_plane_normal}, this optimization produces strong alignment between the Gaussian-wise normal map and the local plane normal map, resulting in a more uniform and coherent surface representation.

\subsection{RF-PGS Representation and Rendering}

\begin{figure}[h]
    \vspace{-3mm}
    \centering
    \includegraphics[width=0.48\textwidth]{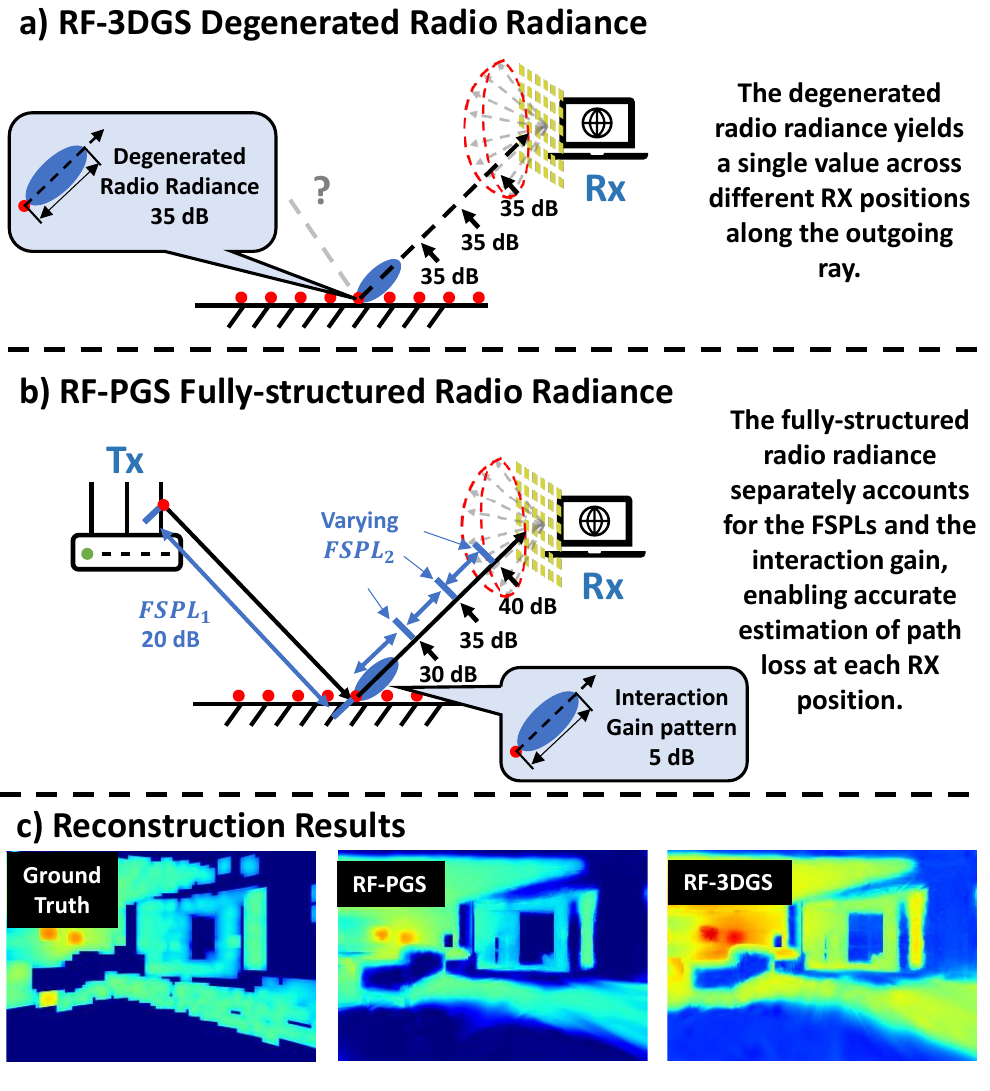} 
    \vspace{-5mm}
    \caption{RF-PGS fully-structured radio radiance.}
    \vspace{-1mm}
    \label{fig:8_RF-PGS_rendering}  
\end{figure}

A key limitation of RF-3DGS is its direct adaptation of visual radiance field methods to the RF domain, which leads to a degenerated radio radiance field, especially in complex indoor environments. As shown in Fig.~\ref{fig:8_RF-PGS_rendering}(c), this results in notable reconstruction errors in areas such as the lobby’s far corner, primarily due to the loose geometric representation that impedes geometry-aware refinement and the extraction of geometry-dependent channel characteristics.
In contrast, RF-PGS leverages accurate geometry to construct a fully-structured radio radiance representation. Its core innovation is the decomposition of each MPC into free-space path loss (FSPL) and interaction gain, enabling direct querying of AoD and ToF after training. This allows RF-PGS to reconstruct physically grounded radio radiance without relying on auxiliary spectra, as demonstrated in Fig.~\ref{fig:3_pipeline_overview} and Fig.~\ref{fig:8_RF-PGS_rendering}(c), where signal attenuation in the lobby’s corner is accurately captured. Furthermore, the recovered radio radiance can be directly aggregated into Tx-side spectra, benefiting downstream tasks. 

In particular, we begin by revisiting how the total path loss of a radio propagation path is computed.
Consider a specific propagation path in environment $\mathcal{E}$ with a known transmitter configuration $\mathcal{T}$. Suppose there are $N$ interactions or retransmission points along the path, denoted as $\mathbf{x_{reTx_i}}$, each contributing an interaction gain $A_i$. The total received-to-transmitted power ratio can be expressed as:
\begin{align}
    \begin{split}
        \frac{P_{\text{Rx}}}{P_{\text{Tx}}} &= 
        \underbrace{G_{\text{Tx}}(\theta_{\text{AoD}}, \phi_{\text{AoD}}) \cdot G_{\text{Rx}}(\theta_{\text{AoA}}, \phi_{\text{AoA}})}_{\text{Antenna Gains}} \\
        &\quad \cdot \underbrace{\prod_{i=1}^{N} A_i}_{\text{Interaction Gains}}
        \cdot \underbrace{\left( \frac{\lambda}{4\pi \sum_{i=1}^{N+1} d_i} \right)^2}_{\text{Free-space Path Loss}},
    \end{split}
    \label{eq::MPC_pathloss}
\end{align}
Here, $\lambda$ denotes the carrier wavelength, and $G_{\text{Tx}}$ is the transmitter antenna gain pattern determined by $\mathcal{T}$. In our implementation, this gain is normalized to an isotropic pattern, consistent with standard channel datasets and measurements~\cite{NIST_data,NIST_SAGE}. The receiver gain pattern $G_{\text{Rx}}$ depends on the scenario. By default, we also assume an isotropic receiver, but for specific applications with directional antennas, angular-domain convolution is sufficient. For array receivers, additional considerations such as array signal processing (ASP) and near-field effects may apply. Beyond these scenario-wise parameters, the AoD, the AoA, and the lengths of individual path segments $d_i$ are all determined by the specific propagation path and should be inherently captured by the learned RF-PGS representation.

However, as shown in Fig.~\ref{fig:8_RF-PGS_rendering}(a), RF-3DGS records only the final segment of the full propagation path. Consequently, all received path loss values ${\frac{P_{Rx}}{P_{Tx}}}$ for receivers located along the outgoing ray  $\mathbf{r}(t) = \mathbf{x_{reTx_N}} + t \mathbf{d_{N+1}}$ are represented by a single radiance channel value $\mathbf{c}_{\mathbf{x_{reTx_N}}, \mathbf{d_{N+1}}}$, where $\mathbf{x_{reTx_N}}$ denotes the last retransmission point and $\mathbf{d_{N+1}}$  is the corresponding outgoing direction. This design implicitly assumes that the length of the final segment $d_{N+1}$ has negligible influence on the received power. This approximation is reasonable in the CV domain, where human visual perception follows an approximately logarithmic response to brightness over a dynamic range exceeding 100 dB. Moreover, visible light typically involve multiple light bounces and long propagation distances (e.g., up to six in standard computer graphics settings), which diminish the perceptual sensitivity to last path segment length $d_{N+1}$.

In the RF domain, dominant propagation paths typically involve fewer bounces and shorter travel distances. In cases such as LoS or single-bounce paths, where the previous path segment (e.g., $d_1$) is short but the final segment ($d_2$) is considerably longer, the resulting path loss deviation can reach tens of dB. As shown in Fig.~\ref{fig:8_RF-PGS_rendering}(a), RF-3DGS fails to account for this variation, assigning a single degenerated value (e.g., 35 dB) to all receivers along the outgoing ray, regardless of their distance from the retransmission point. This averaging effect leads to substantial prediction errors at unseen locations, particularly when the final segment dominates the total path length, as illustrated in Fig.~\ref{fig:8_RF-PGS_rendering}(c).
Moreover, RF-3DGS lacks access to preceding path segments, preventing accurate reconstruction of ToF and AoD. Since the representation is tied to the AoA domain at the receiver, leveraging the full channel information at the Tx-side requires additional transformation from AoA to AoD domain, complicating its use in downstream tasks.

Thus, in RF-PGS, we explicitly record a fully-structured radio radiance field. In our current experimental setup in an indoor lobby environment of approximately $14 m \times 15 m$ area, dominant signal components are primarily single-bounce paths. This observation is supported by both the digital twin framework used in RF-3DGS~\cite{zhang2024rf} and our previously collected indoor spatial channel dataset~\cite{zhang2024wisegrt}.
To accommodate this, the RF-PGS radiance representation is designed to be decomposed into two components: the FSPL, which depends on the total path length, and the interaction gain pattern, which arises from surface reflection or diffraction. As shown in Fig.~\ref{fig:8_RF-PGS_rendering}(b), for a given query ray from a receiver position into the RF-PGS model, the ray–surface intersection point is first estimated. Based on this, the distances to both the transmitter and receiver can be accurately computed, allowing precise calculation of the FSPL term in Eq.~\ref{eq::MPC_pathloss}.
Crucially, the SH function stored on each intersected Gaussian no longer encodes degenerated radio radiance, but instead represents a fixed interaction gain pattern that captures the intrinsic surface scattering behavior, as shown in Fig.~\ref{fig:8_RF-PGS_rendering}(b). Given that multiple overlapping planar Gaussians may contribute to a single query ray, we adopt the same differentiable alpha-blending strategy used in RF-3DGS to aggregate interaction gains. As a result, reconstruction remains fully differentiable and efficient. Unlike RF-3DGS, however, RF-PGS calibrates the FSPL component using precise Tx–Rx geometry, ensuring that the final radiance prediction better aligns with ground truth.

This decomposition further enables RF-PGS to recover directional interaction patterns that reflect the underlying radio-surface system function, facilitating downstream tasks such as RF parameters calibration of digital twin. However, in more general large-scale outdoor environments, strong multi-bounce propagation paths may play a critical role, particularly in non-line-of-sight (NLoS) regions. While RF-PGS currently focuses on single-bounce interactions, it can help identify potential multi-bounce paths by detecting the deviations from expected specular reflection—such as when the angle between incident and outgoing rays differs significantly from the local surface normal, or when the last retransmission point is not the transmitter but an intermediate scatterer. To fully model these multi-bounce paths, RF-PGS can be integrated with a digital twin framework using ray tracing, or extended with a dedicated multi-bounce path finding algorithm.

\subsection{Multi-view Loss for Handling Imperfections in Radio Spatial Spectra}
To address imperfections in radio spatial spectra, we propose a multi-view loss strategy. These inconsistencies primarily arise during the ASP stage, where various types of spectra exhibit different forms of discrepancy. Consequently, the loss function must be adapted to account for these variations.
Generally, the multi-view loss we introduce is designed to handle inter-view inconsistencies. For each view, we select a set of nearest views based on two criteria: (1) the angular difference between view directions must fall below a predefined threshold, and (2) the Euclidean distance between the corresponding Rx positions must be within a specific range. During training, one of these nearest views is randomly selected to compute the multi-view loss. More specifically, for different spectra, we utilize two types of multi-view loss.

For beamforming techniques such as conventional beamforming (CBF) and tapered CBF (TCBF), the spectrum is constructed by querying signals along specific AoAs, where reception is achieved by forming a main lobe in the target direction, accompanied by side lobes that may introduce interference. Although large arrays employing side-lobe suppression or adaptive beamforming can mitigate side-lobe effects, this typically comes at the cost of a wider main lobe. As shown in Fig.~\ref{fig:3_pipeline_overview}, a wider main lobe spreads the energy of the strong MPC to surrounding query directions, causing the corresponding surrounding Gaussians to learn the MPC in incorrect directions. As a result, the directional interaction pattern is blurred or disrupted.

In contrast, neighboring views (such as the upper radio spatial spectra in Fig.~\ref{fig:3_pipeline_overview}) can observe the same Gaussians without such distortion, thus providing cleaner supervision signals with only minor deviations due to the small angular differences between the views. For each query direction, we extract an adaptive-size patch of its surrounding directions, obtaining their estimated radio-surface intersections in world coordinates. These intersections are then projected into the camera coordinate of the selected nearest view, and the corresponding pixel values are retrieved. A minimum-merge operation is applied to synthesize a reference patch, and the normalized cross-correlation loss is computed between this reference patch and the original patch. This approach significantly reduces the interference in beamforming spectra.

For MPC estimation methods, such as SAGE~\cite{NIST_SAGE} and RIMAX~\cite{thoma2004rimax}, inconsistencies arise from the unstable recovery of weaker MPCs across adjacent samples. This is primarily due to the limited number of components allowed in each estimation. While strong MPCs are consistently estimated, weaker MPCs may intermittently appear or vanish across samples. To mitigate this, our loss encourages consistency for weaker MPCs across neighboring views. Similar to the beamforming case, we extract patches across views, but in this scenario, we apply a maximum-merge operation instead of minimum-merge. This ensures that weaker MPCs, which may be lost in individual views, are preserved.

\section{Experimental Results}
\subsection{Experimental Setup and Evaluation}
\begin{figure}[h]
    \centering
    \vspace{-3mm}
    \includegraphics[width=0.45\textwidth]{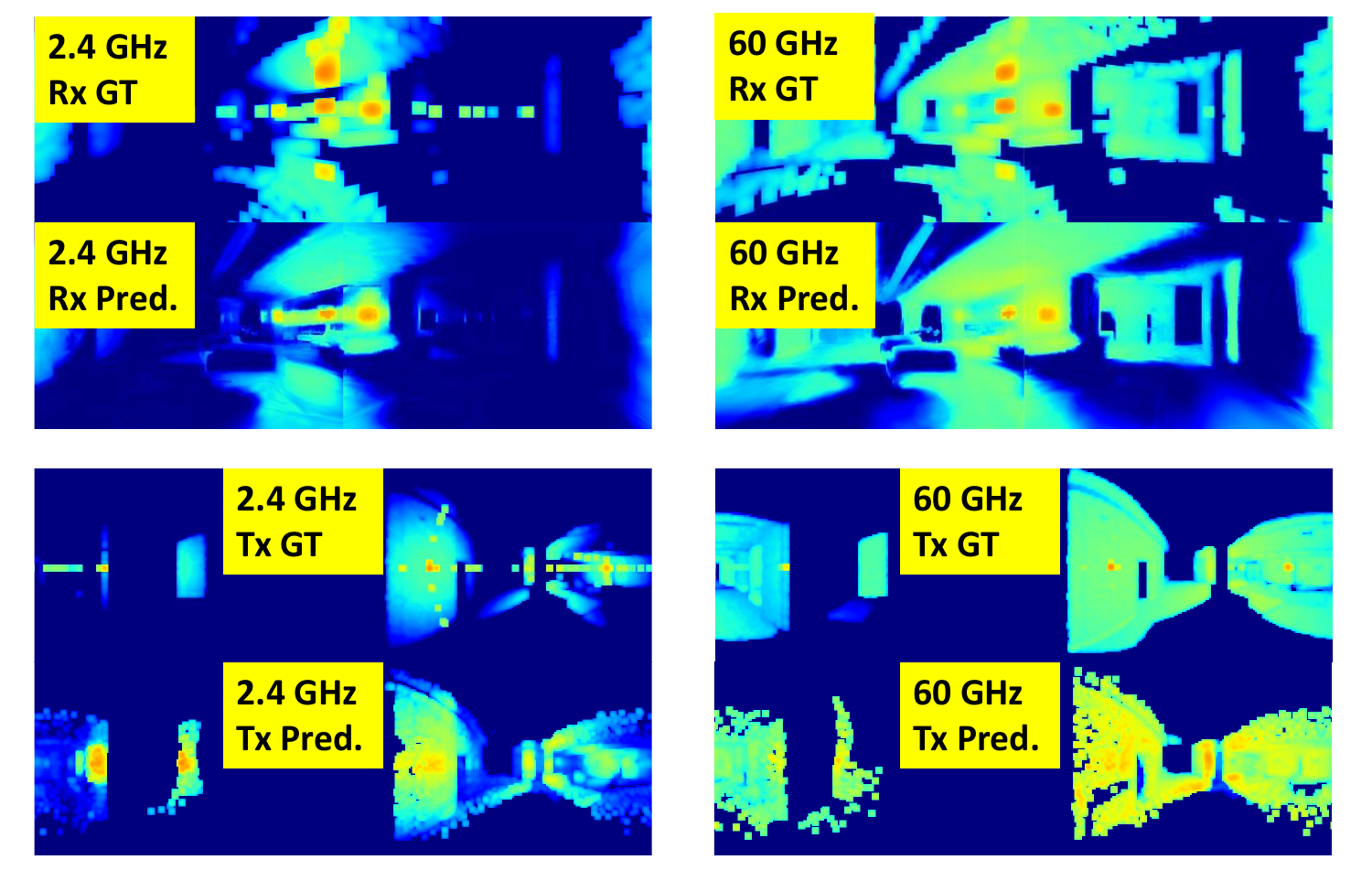}
    \vspace{-3mm}
    \caption{RF-PGS predictions (Pred.) compared to ground truths (GT).}
    \vspace{-3mm}
    \label{fig:9_result_demo}  
\end{figure}

In this work, the proposed RF-PGS framework and all baseline methods are implemented on a desktop equipped with an AMD Ryzen 7 7700 CPU and an RTX 4090 GPU. Experiments are conducted on both simulated and real-world field measurement datasets.
For simulation, we use Sionna~\cite{hoydis2023sionna}, an open-source wireless simulator developed by NVIDIA that integrates a ray-tracing engine and supports GPU acceleration. Two frequency bands are simulated: 2.4~GHz and 60~GHz, with material properties carefully configured to match real-world physical characteristics.
These two bands exhibit distinct propagation behaviors, as illustrated in Fig.~\ref{fig:9_result_demo}. At 2.4~GHz, radio waves show stronger specular reflection and diffraction, whereas at 60~GHz, scattering effects are more pronounced. To better highlight strong signal paths, we apply path loss thresholds of $-160$~dB for 2.4~GHz and $-190$~dB for 60~GHz. This contrast-enhancing strategy allows better focus on dominant propagation paths, though it introduces additional challenges and leads to a slight performance drop for RF-3DGS. 

For evaluation, 100 test samples are randomly selected from each dataset and kept fixed across all training configurations. Training sets are constructed with varying sizes: 10, 15, 20, 25, 50, 100, 200, 400, and 700 samples. After training, the reconstructed radio spatial spectra at the same test poses are compared to those generated from the ground-truth test data.
Since RF-PGS supports direct querying from both transmitter and receiver perspectives, we evaluate the reconstructed spectra quality at both the Rx and Tx sides, providing a more comprehensive assessment.

To evaluate RF-PGS in more realistic conditions, we also test it on 60~GHz field measurement data, which is provided by the National Institute of Standards and Technology (NIST), USA~\cite{NIST_data}. The dataset contains the full Spatial-CSI of estimated MPCs with SAGE algorithm and is further processed into the path loss spectra at both Tx and Rx sides.

\subsection{General Comparison with Prior Methods}
We begin with a general comparison between our proposed method, RF-PGS, and several representative baseline methods. The first baseline is RF-3DGS, which leverages 3D Gaussian Splatting but lacks accurate geometry representation and full-path modeling. The second baseline is NeRF$^2$, a neural radiance field-based method adapted for RF scenarios, representing neural radiance field methods. Lastly, we include conditional GAN (CGAN) method as a representative of data-driven generative models. Table~\ref{tab:general_comparison} summarizes the average performance of each method across several key metrics, including Peak Signal-to-Noise Ratio (PSNR, the higher the better), Structural Similarity Index Measure (SSIM, the higher the better), and Learned Perceptual Image Patch Similarity (LPIPS, the lower the better), evaluated on Rx-side spectra. Additionally, we report the average training time and inference time to assess computational efficiency.

\begin{table}[h]
    \centering
    \caption{General performance comparison.}
    \label{tab:general_comparison}
    \begin{tabular}{lcccccc}
        \toprule
        \textbf{Method} & \textbf{PSNR} $\uparrow$ & \textbf{SSIM} $\uparrow$ & \textbf{LPIPS} $\downarrow$ & \textbf{Train} $\downarrow$ & \textbf{Infer.} $\downarrow$  \\
        \midrule
        RF-PGS     & \textbf{20.6108} & \textbf{0.6606} & \textbf{0.3945} & 3min~53s & 4~ms  \\
        RF-3DGS    & 14.2208 & 0.3680 & 0.4250 & \textbf{2min~41s} & \textbf{2~ms} \\
        NeRF$^2$   & 14.1335 & 0.3556 & 0.7031 & 2h~32min & 0.7~s  \\
        CGAN       & 10.1030 & 0.1832 & 0.8912 & 1h~27min & 5~ms \\
        \bottomrule
    \end{tabular}
\end{table}
As shown in the results, RF-PGS outperforms the baseline methods across evaluation metrics. It achieves much higher reconstruction quality, indicated by superior PSNR, SSIM, and LPIPS scores, while also demonstrating significantly shorter training and inference times compared to NeRF$^2$ and CGAN. Notably, as shown in our previous work~\cite{zhang2024rf}, NeRF$^2$ fails to reconstruct meaningful geometry when using only RF path loss spectra as input under its default configuration, often producing vague floating artifacts. While NeRF$^2$ can also be adapted to use visual data for geometry reconstruction, freezing the density MLP, and then performing RF-training on the feature MLP, its visual training stage alone typically requires over 24 hours to achieve geometry quality comparable to that of RF-PGS. Moreover, prior neural radiance field and 3DGS-based methods rarely support direct and simultaneous rendering of both Tx-side and Rx-side spectra. Thus, among all baselines, RF-3DGS is the only method that exhibits capability and performance close to RF-PGS. Therefore, in the following sections, we focus on detailed comparisons between RF-PGS and RF-3DGS to further analyze the specific improvements introduced by our proposed enhancements.

\begin{figure}[h]
    \centering
    \includegraphics[width=0.48\textwidth]{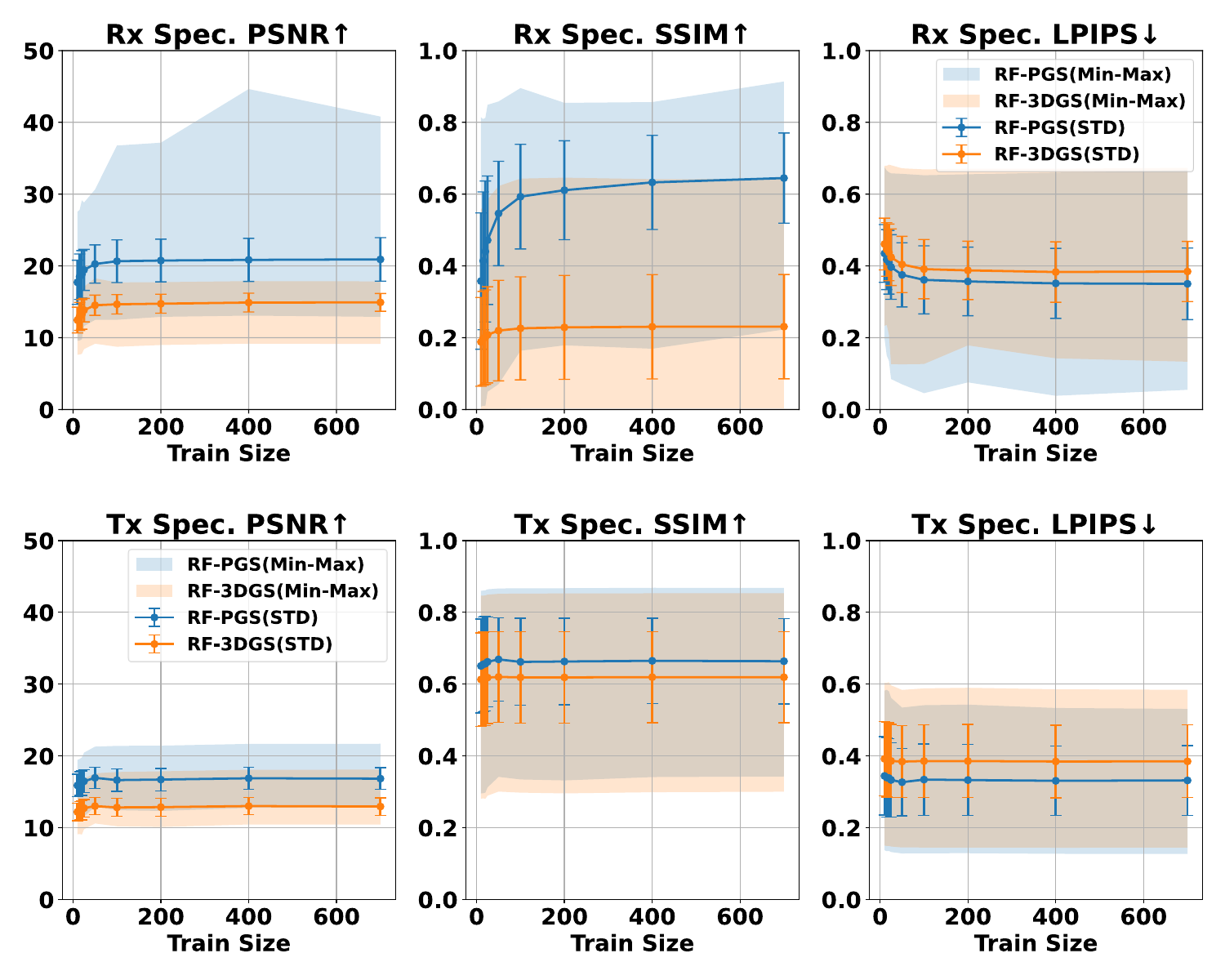}
    \caption{2.4~GHz reconstruction evaluation. Upper: similarity evaluation of the queried Rx-side spectra to the ground truth. Lower: evaluation of the Tx-side spectra, which is more challenging since Tx-side spectra are not included in the training dataset and rely solely on the model’s ability to capture the underlying physical radio propagation.}
    \vspace{-5mm}
    \label{fig:10_24GHzRxmetric}  
\end{figure}

\subsection{Performance Analysis in Comparison to RF-3DGS}
\subsubsection{2.4~GHz Simulation Dataset Reconstruction Results.}
In Fig.~\ref{fig:10_24GHzRxmetric}, we show the reconstruction quality of both Rx-side and Tx-side spectra using PSNR, SSIM, and LPIPS metrics under varying training set sizes. First, regarding the training set size, we observe that RF-PGS (blue line and area) is insensitive to the amount of training data. Performance remains stable across a wide range of sample sizes, with a noticeable drop only when the number of samples falls below 20.
For Rx-side spectra, RF-PGS achieves significantly better performance than RF-3DGS (orange line and area) in terms of both PSNR and SSIM. This holds not only for average values but also for the minimum and maximum values across all test groups. The superior performance of RF-PGS is mainly attributed to its explicit modeling of the full radio propagation path, which enables accurate path loss querying at arbitrary Rx poses. In contrast, RF-3DGS can only learn an average approximation of the path loss from training samples. 
However, in terms of LPIPS which captures high-level perceptual similarity, RF-PGS shows less improvement over RF-3DGS compared to the PSNR and SSIM. The reason is RF-3DGS already reconstructs the spatial structure reasonably well, but only lacks numerical accuracy. The observed LPIPS improvements in RF-PGS can be attributed to its denser geometry, more accurate diffraction handling, and the incorporation of multi-view loss.

As for the Tx-side spectra, it is important to note that the original RF-3DGS cannot directly render such spectra, but relies on rendering the AoD spectrum and path loss, and then transforming them to Tx-side spectra, which incurs extra computational overhead. Moreover, Tx-side spectrum reconstruction is inherently more challenging than Rx-side. This is because the training input consists only of Rx-side spectra; hence, the training process primarily matches the Rx views. The Tx-side spectra are derived from the reconstructed radio radiance field, and their fidelity depends entirely on whether the representation accurately captures the underlying physical reality. The high-quality Tx-side reconstruction achieved by RF-PGS further validates its capability to learn true radio propagation behaviors. 

\begin{figure}[h]
    \centering
    \includegraphics[width=0.48\textwidth]{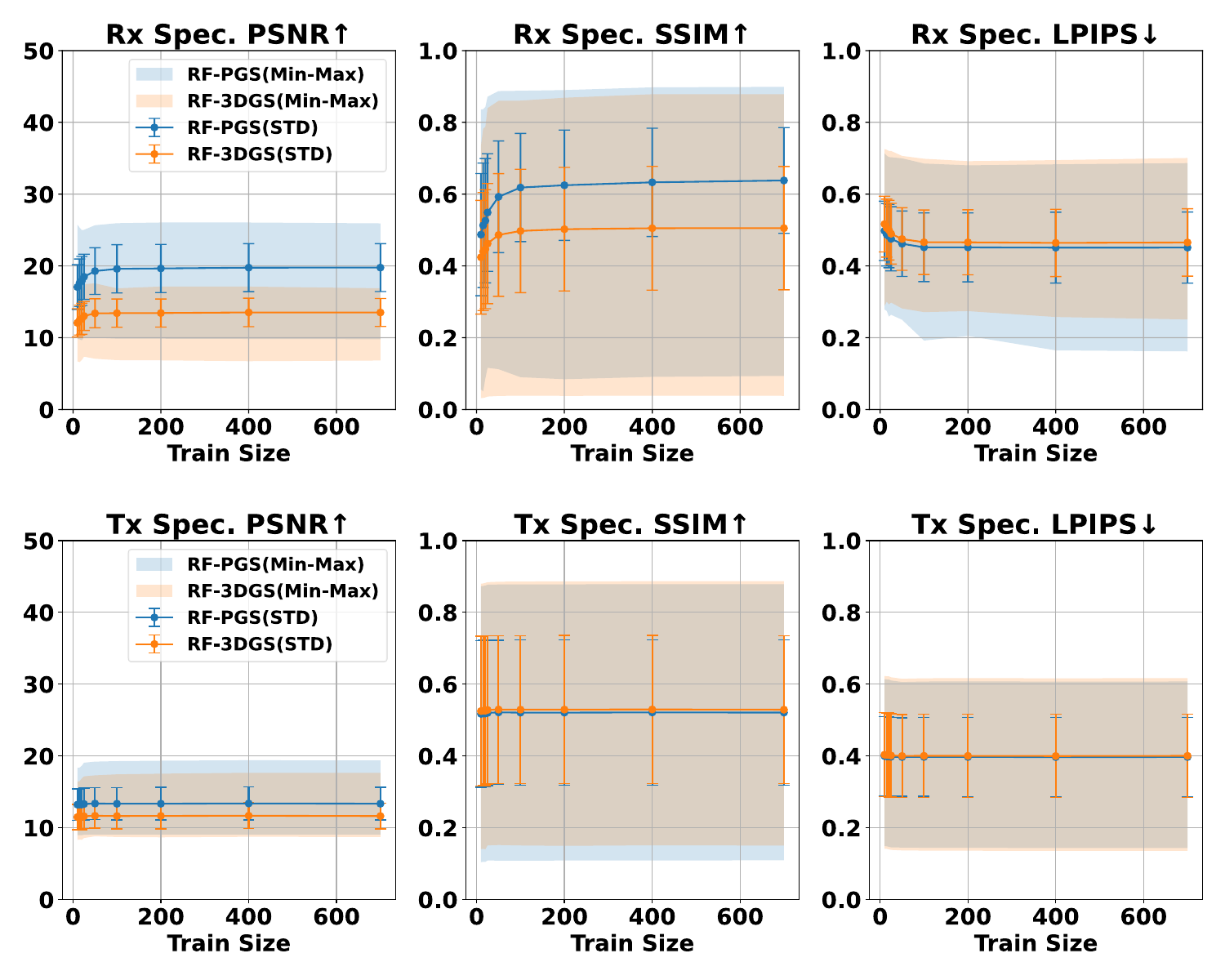}
   
    \caption{60~GHz reconstruction evaluation.}

    \label{fig:11_60GHzRxmetric}  
\end{figure}
\subsubsection{60~GHz Simulation Dataset Reconstruction Results.}
For the 60~GHz scenario, RF-PGS continues to outperform RF-3DGS on the Rx-side spectra in all three metrics. Notably, RF-3DGS demonstrates better performance at 60~GHz compared to the 2.4~GHz dataset. This is largely due to the higher prevalence of scattering paths at 60~GHz, which contributes more significantly to the evaluation metrics and RF-3DGS is good at reconstructing such structural information.

For Tx-side spectrum reconstruction at 60~GHz, the performance improvement offered by RF-PGS is further limited. PSNR and SSIM show only marginal gains, while the LPIPS scores are nearly unchanged. This limitation primarily results from the difficulty of accurately reconstructing the large number of scattering paths, which do not follow the reflection principle, and the Tx-side spectrum relies entirely on querying from the reconstructed full propagation paths. However, in practical wireless communication scenarios, the primary concern is often the accurate reconstruction of dominant paths. For sensing-related applications, structural fidelity tends to be more important than numerical precision. Nevertheless, improving the modeling and estimation of scattering effect remains an open challenge for future research.

\subsubsection{Spatial Domain Beamforming Test with Simulation Dataset}
\begin{figure}[ht]
    \centering
  
    \includegraphics[width=0.4\textwidth]{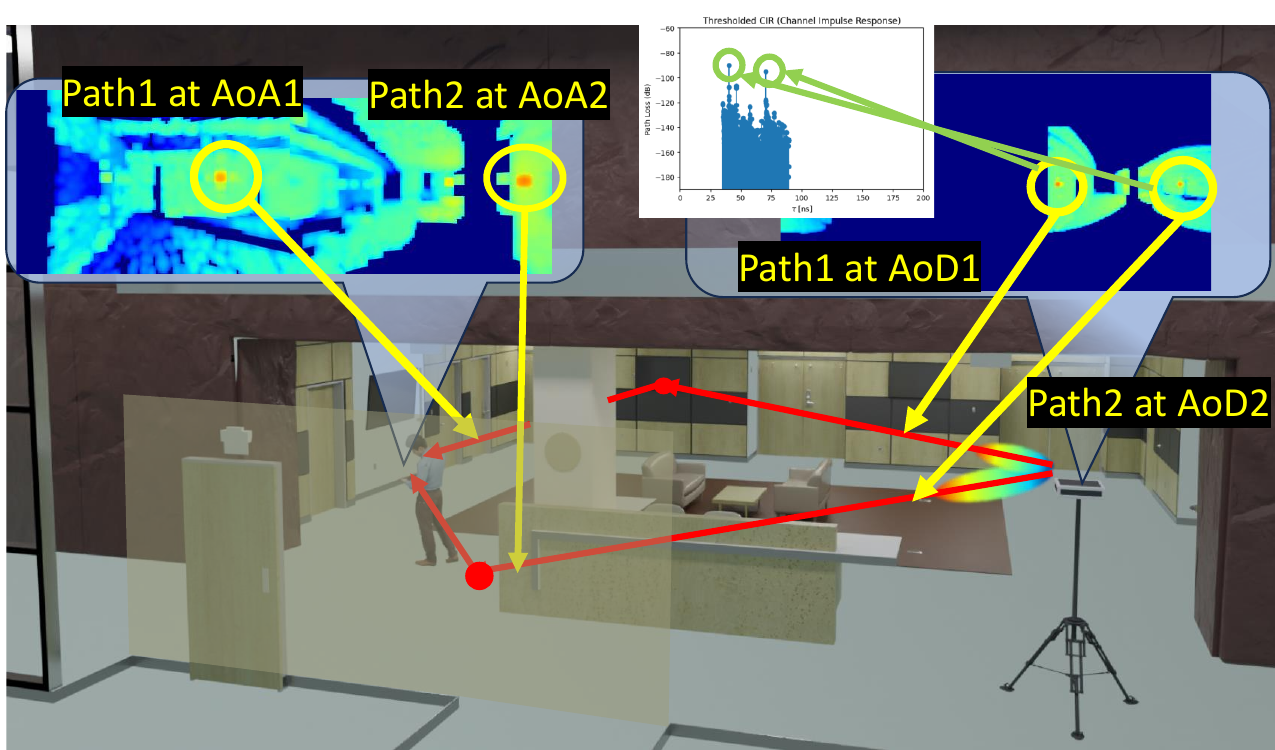} 
    
    \caption{Illustration of spatial domain beamforming concept.}

    \label{fig:13_angle-space-beamforming}  
\end{figure}

One important application of RF-PGS is enabling spatial domain beamforming in wireless communication systems. Since the spatial structure of the wireless channel remains relatively stable with respect to the environment, explicitly modeling the environment-specific spatial channel emerges as a promising alternative to traditional estimation approaches. With a well-trained RF-PGS model, the beamforming process can be significantly simplified. As illustrated in Fig.~\ref{fig:13_angle-space-beamforming}, for each downlink transmission, the base station (BS) only needs to steer main lobes along the known propagation paths toward the target user while suppressing interference toward non-target users. This strategy not only reduces reliance on dense pilot signals but also facilitates efficient hybrid beamforming with minimal calibration requirements.
Additionally, pilot overhead for baseband channel equalization can be substantially reduced. For example, delay-alignment modulation~\cite{lu2022DAm} can now be explicitly applied at the BS, resulting in a clear single-impulse response at the receiver. This capability can also benefit various emerging 6G technologies, such as reconfigurable intelligent surfaces (RIS)~\cite{liu2021reconfigurable} and integrated sensing and communication (ISAC)~\cite{isac}.

\begin{figure}[h]
    \centering
    \includegraphics[width=0.48\textwidth]{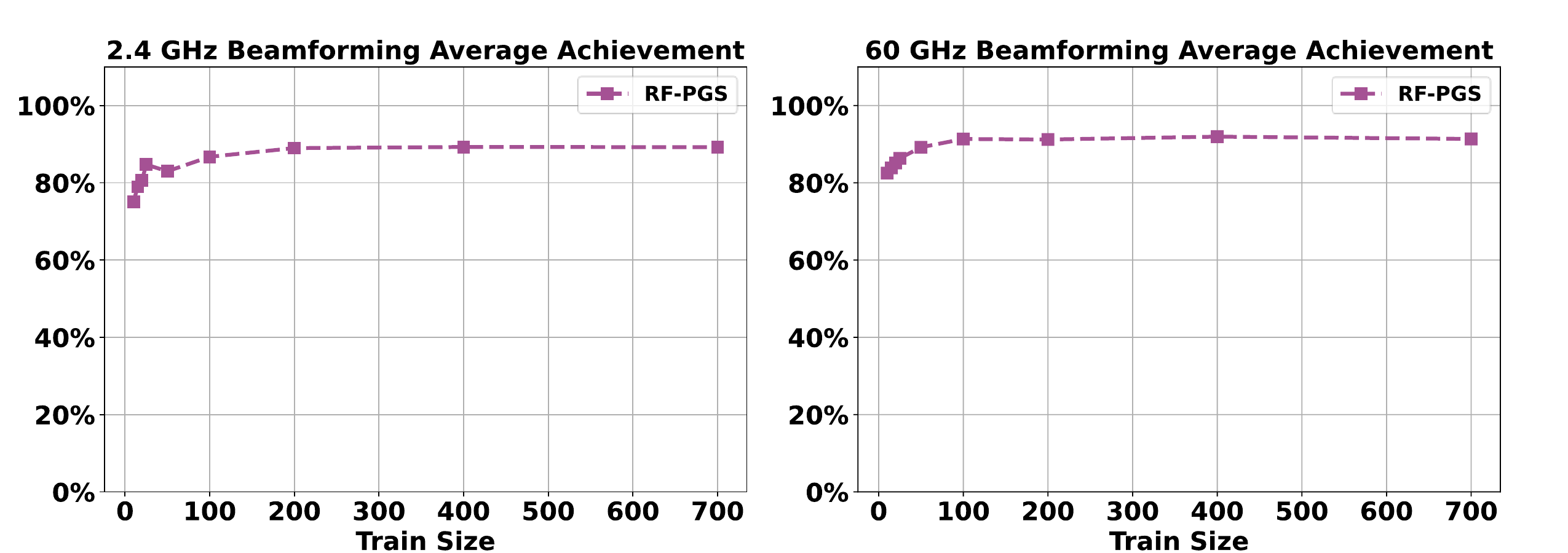}
    \caption{Beamforming channel capacity on simulation datasets.}

    \label{fig:12_reconstruction_beamforming_rate}  
\end{figure}

To evaluate the effectiveness of RF-PGS in such scenarios, we test its spatial domain beamforming performance using simulation datasets. For both 2.4~GHz and 60~GHz setups, we configure nearly identical system parameters. The only distinction is that the 60~GHz scenario employs larger antenna arrays to maintain the same effective aperture. As a result, the theoretical channel capacities are nearly equivalent, allowing us to isolate and analyze the impact of different propagation physics across the two frequency bands. In both cases, beamforming at the transmitter and receiver is performed using only the reconstructed path loss spectra.

As shown in Fig.~\ref{fig:12_reconstruction_beamforming_rate}, the achieved communication rates in both scenarios closely approach the theoretical capacity. This is primarily due to the fact that the beam patterns derived from our reconstructed Tx and Rx spectra are well aligned with the actual physical propagation paths. More importantly, even under extremely sparse supervision with only ten training samples, the system still achieves robust performance (up to 92\% of optimal capacity) with minimal degradation in data rate.

\begin{figure}[h]
    \centering
    \includegraphics[width=0.48\textwidth]{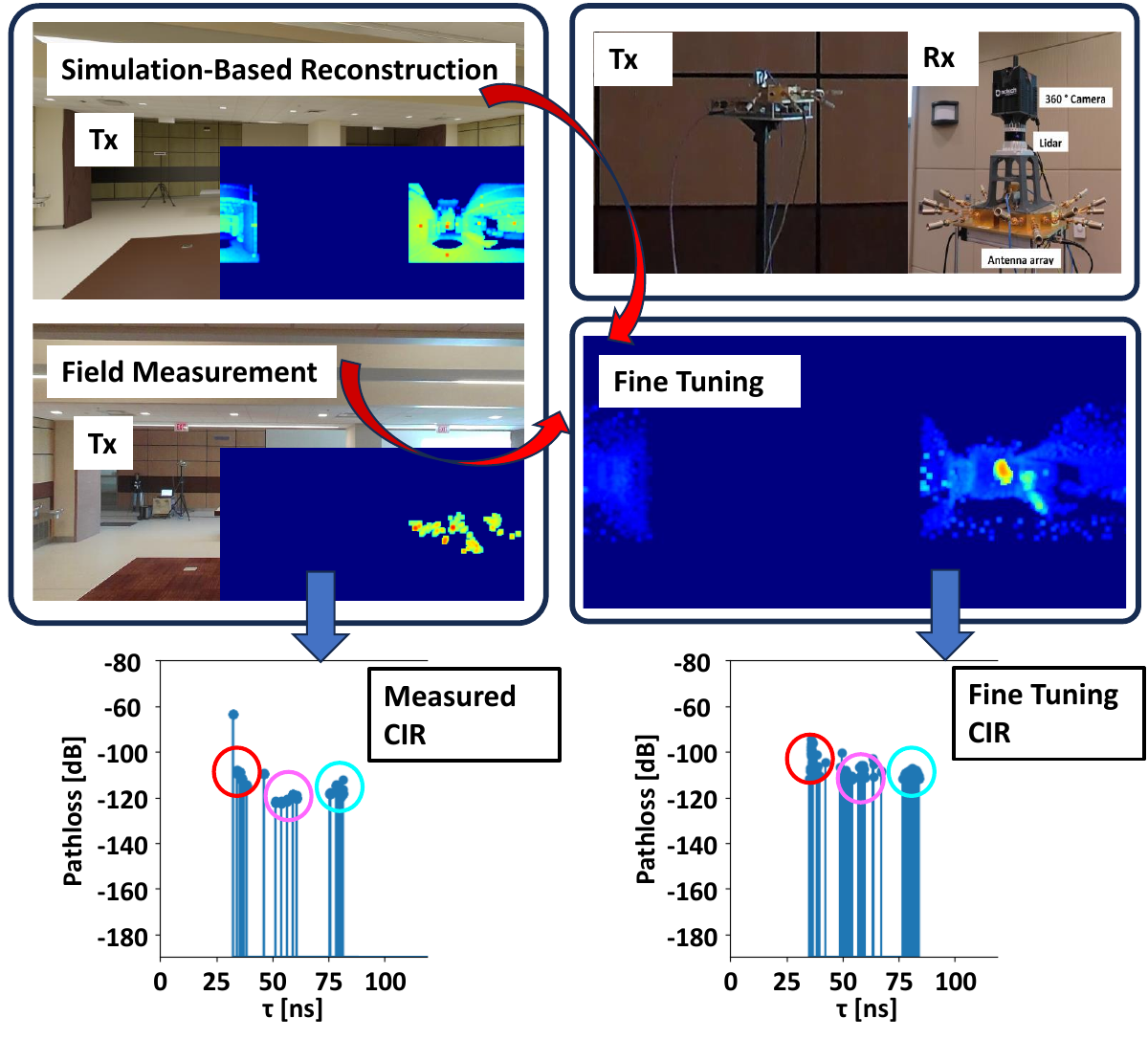}
    \vspace{-5mm}
    \caption{Demonstration of simulation, field measurement, and fine-tuned reconstruction. }
    \vspace{-5mm}
 
    \label{fig:14_field_measurement}  
\end{figure}

\begin{figure}[h]
    \centering
    \includegraphics[width=0.5\textwidth]{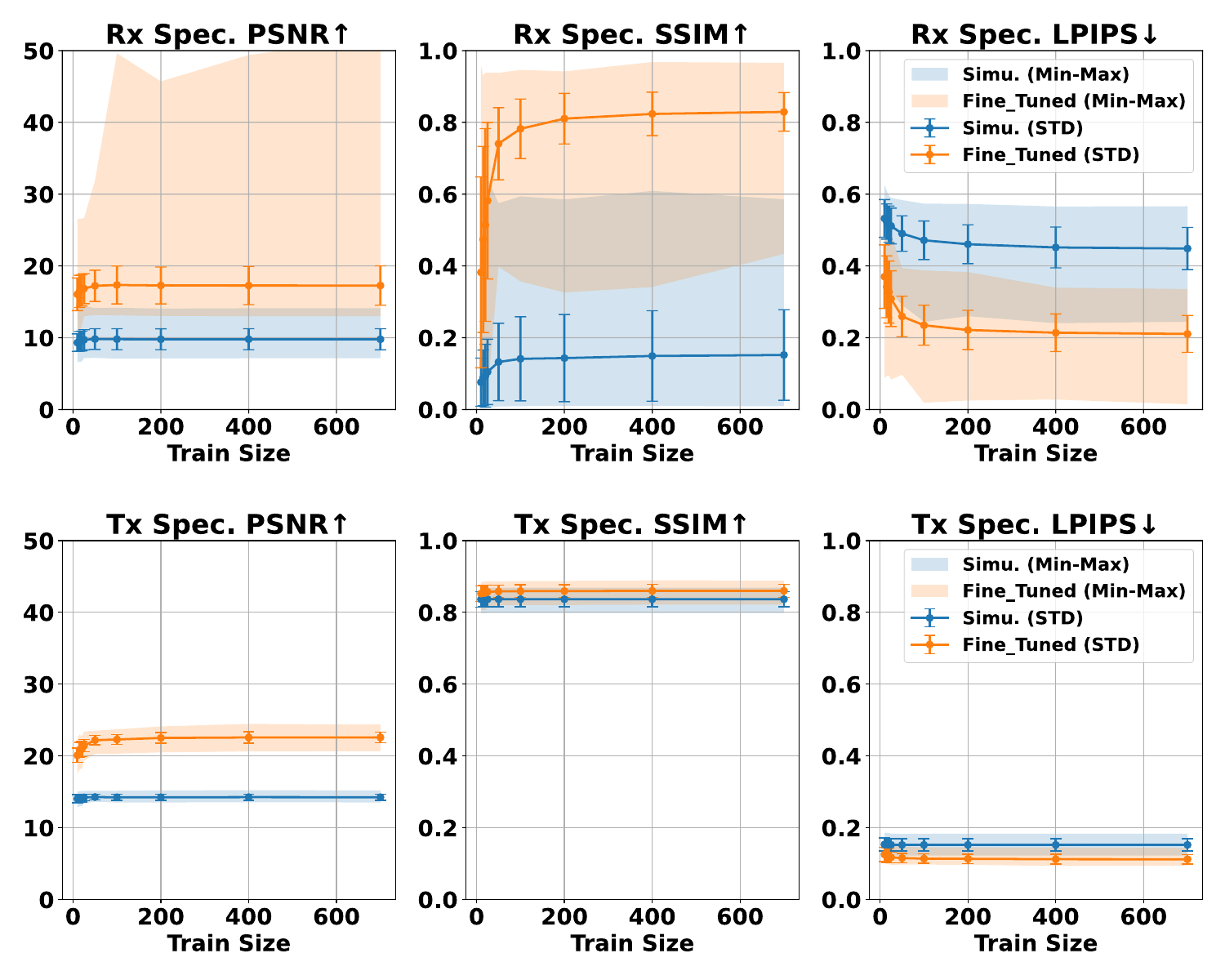}
    \caption{Validation of simulation and fine-tuning results.}
    \label{fig:15_NIST_finetuning}

\end{figure}

\begin{figure}[h]
    \centering
    \includegraphics[width=0.48\textwidth]{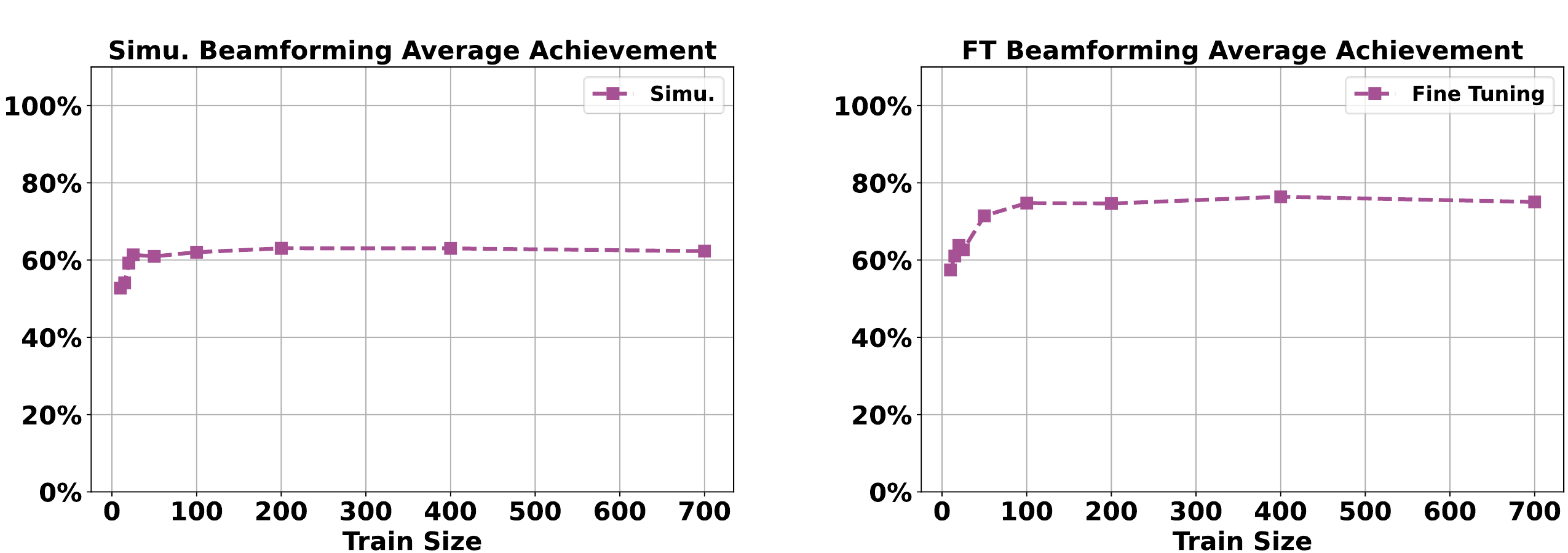}
    \caption{Achieved average beamforming rate at test positions. Left: reconstruction based solely on simulation data for spatial domain beamforming in the field measurement scenario. Right: fine-tuned reconstruction using field measurement data to guide spatial domain beamforming.}
    \label{fig:16_colocated_beamforming_rate}  
\end{figure}

\vspace{-4mm}
\subsection{Validation of Simulation Pipeline and Fine-tuning with Measurement Data.}
To validate our digital simulation pipeline against field measurement and to evaluate RF-PGS in a realistic scenario, we adopt a digital-twin-style framework. Field measurements in the same lobby were conducted by NIST USA; details of the campaign are provided in \cite{NIST_data,NIST_SAGE}. In this setup, the Tx has a field-of-view (FoV) of $180^\circ$ in azimuth and $90^\circ$ in zenith, and the Rx has a FoV of $360^\circ$ in azimuth and $90^\circ$ in zenith. The antenna array directional gain is normalized to 1 in all directions, and MPCs are estimated by the SAGE algorithm, which focuses on recovering tens of dominant MPCs. We replicate this configuration by constructing a simulation scene with the same Tx/Rx FoVs and sampling positions, as shown in the upper-left panel of Fig.~\ref{fig:14_field_measurement}. Comparing the Tx-side spectra from the simulation-based reconstruction with the measured MPCs unfolded in the AoD domain shows that the SAGE-processed MPCs are significantly sparser than the simulation results with discrepancies in path loss. To quantify this gap, we first directly compare the simulation-based reconstruction with the field measurements.

As shown by the blue lines and area in Fig.~\ref{fig:15_NIST_finetuning}, although the LPIPS values are not particularly poor, the statistical PSNR and SSIM values are very low, due to the discrepancies in material properties and interaction parameters. Fortunately, one major advantage of RF-PGS is its rapid training capability, which allows efficient fine-tuning using a small amount of real-world data. Specifically, depending on the size of the fine-tuning set, we limit training to either 500 iterations or the full fine-tuning set, whichever is smaller, resulting in a fine-tuning duration of less than 30 seconds. Then, we apply adaptive learning rates and incorporate the proposed max-merge multi-view loss during training. After fine-tuning (Fig.~\ref{fig:15_NIST_finetuning}), both PSNR and SSIM improve substantially. An example in the lower part of Fig.~\ref{fig:14_field_measurement} visualizes the measured CIR and the fine-tuned CIR, where the fine-tuned RF-PGS model accurately captures the dominant MPC clusters (at 35, 50, and 80 ns) in this complex environment.

We also conduct beamforming tests under the same setup as before. When directly applying the simulation-based model to the field measurement environment (using the SAGE-processed MPCs as ground truth), the communication rate remains reasonably high due to the accuracy of the geometry of the strong paths. After fine-tuning, the rate improves further, although it remains slightly lower than in the fully simulated scenario. This is primarily because the ground truth channel, the SAGE-estimated MPCs as shown in Fig.~\ref{fig:14_field_measurement} are sparser than the beamforming-processed simulation spectra.

\section{Conclusion}
In this work, we proposed RF-PGS, a novel radiance field reconstruction framework. By incorporating accurate planar geometry, a hierarchical inverse tracing algorithm, and further optimization, RF-PGS enables high-fidelity reconstruction of full radio propagation path using only Rx-side path loss spectra. Extensive experiments on both simulated and real-world datasets demonstrate its superiority in reconstruction accuracy, computational efficiency, and robustness under sparse supervision.

Despite these advancements, RF-PGS still has several limitations. The learned representation remains site-specific and cannot generalize across environments with similar features but different layouts and materials. Additionally, the current implementation is limited to static scenes and does not account for dynamic elements such as moving objects or users. Addressing these limitations would enhance the applicability of RF-PGS in more realistic and complex wireless communication scenarios.

\bibliographystyle{ieeetr}
\bibliography{sample-base}

\end{document}